\documentclass[10pt,journal,compsoc]{IEEEtran}
\ifCLASSOPTIONcompsoc
  \usepackage[nocompress]{cite}
\else
  \usepackage{cite}
\fi
\ifCLASSINFOpdf
\else
\fi
\usepackage[utf8]{inputenc} % allow utf-8 input
\usepackage[T1]{fontenc}    % use 8-bit T1 fonts
\usepackage{hyperref}       % hyperlinks
\usepackage{url}            % simple URL typesetting
\usepackage{booktabs}       % professional-quality tables
\usepackage{amsfonts}       % blackboard math symbols
\usepackage{nicefrac}       % compact symbols for 1/2, etc.
\usepackage{microtype}      % microtypography
\usepackage{amsmath,amssymb,amsfonts}
\usepackage{algorithmic}
\usepackage{graphicx}
\usepackage{textcomp}
\usepackage{xcolor}
\usepackage{amsthm}
\usepackage{enumerate}
\usepackage{multirow}
\usepackage{caption}
\usepackage{subfigure}
\usepackage{balance}
\usepackage{bbm}
\usepackage{array}
\usepackage{tabularx}
\usepackage{tabu}
\usepackage{wrapfig}
\usepackage{cite}
\newtheorem{theorem}{Theorem}

\begin{document}
\title{Maximum Density Divergence \\for Domain Adaptation}

\author{Jingjing Li, Erpeng Chen, Zhengming Ding, Lei Zhu, Ke Lu and Heng Tao Shen
\IEEEcompsocitemizethanks{\IEEEcompsocthanksitem Jingjing Li, Erpeng Chen, Ke Lu and Heng Tao Shen are with the School of Computer Science and Engineering, University of Electronic Science and Technology of China; Lei Zhu is with Shandong Normal University; Zhengming Ding is with Department of Computer, Information and Technology, Indiana University-Purdue University Indianapolis. Email: lijin117@yeah.net }}

\markboth{IEEE Transactions on Pattern Analysis and Machine Intelligence}%
{Jingjing Li \MakeLowercase{\textit{et al.}}: Maximum Density Divergence for Domain Adaptation}


\IEEEtitleabstractindextext{%
\begin{abstract}
 Unsupervised domain adaptation addresses the problem of transferring knowledge from a well-labeled source domain to an unlabeled target domain where the two domains have distinctive data distributions. Thus, the essence of domain adaptation is to mitigate the distribution divergence between the two domains. The state-of-the-art methods practice this very idea by either conducting adversarial training or minimizing a metric which defines the distribution gaps. In this paper, we propose a new domain adaptation method named Adversarial Tight Match (ATM) which enjoys the benefits of both adversarial training and metric learning. Specifically, at first, we propose a novel distance loss, named Maximum Density Divergence (MDD), to quantify the distribution divergence. MDD minimizes the inter-domain divergence (``match'' in ATM) and maximizes the intra-class density (``tight'' in ATM). Then, to address the equilibrium challenge issue in adversarial domain adaptation, we consider leveraging the proposed MDD into adversarial domain adaptation framework. At last, we tailor the proposed MDD as a practical learning loss and report our ATM. Both empirical evaluation and theoretical analysis are reported to verify the effectiveness of the proposed method. The experimental results on four benchmarks, both classical and large-scale, show that our method is able to achieve new state-of-the-art performance on most evaluations. Codes and datasets used in this paper are available at {\it github.com/lijin118/ATM}. % which has not been explored before
\end{abstract}

}

\maketitle

\IEEEdisplaynontitleabstractindextext

% \fi
\IEEEpeerreviewmaketitle

\IEEEraisesectionheading{\section{Introduction}\label{sec:introduction}}

\IEEEPARstart{T}{he} sweeping success of deep learning in the past decade is inseparable from massive labeled training data, e.g., ImageNet~\cite{ILSVRC15}. However, most of existing large-scale labeled datasets~are~proposed for general tasks. Conventional deep models trained on these datasets cannot handle novel instances, unseen categories, and ad hoc applications~\cite{pan2010survey,hoffman2018cycada,long2018conditional}. In other words, they are delicate in generalizing to unexpected tasks. A slight change in the training domain can lead to destructive result variation in the test domain~\cite{ganin2014unsupervised,hoffman2018cycada,shao2014generalized}. Many real-world applications, however, have to be robust to novel instances with limited or even no labeled training samples available. Recently, domain adaptation has been proven effective in generalizing learned knowledge to novel tasks~and~new~environments~\cite{tzeng2017adversarial,hoffman2018cycada,long2018conditional,li2020deep}. 

A typical domain adaptation setting consists of a well-labeled source domain and an unlabeled target domain~\cite{pan2010survey,kouw2019review,li2018transfer}. In general, the two domains are interconnected, e.g., sharing the same semantic space, but drawn from distinctive data distributions, e.g., street scenes from real cameras and video games~\cite{hoffman2018cycada,venkateswara2017Deep,ding2018robust}. Thus, the main challenge of domain adaptation is how to effectively mitigate the distribution shift between domains. 

\begin{figure}[t]
\begin{center}
\includegraphics[width=0.81\linewidth]{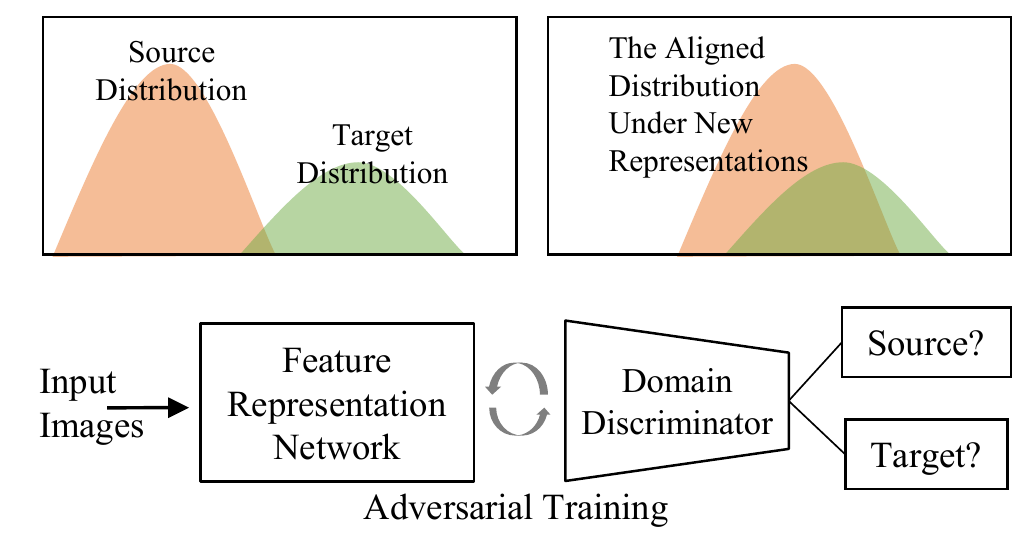}
\end{center}
\vspace{-13pt}
\caption{\small Illustration of domain adaptation and adversarial domain adaptation networks (ADAN). In domain adaptation tasks, the source domain and the target domain have different data distributions. The goal is to learn a new feature representation where the two domains can be well aligned. ADAN leverages the idea of adversarial learning~\cite{goodfellow2014generative} and it assumes that the two domains are aligned as long as the domain discriminator is confused. However, recent advances reveal that such an assumption may be not solid~\cite{arora2017generalization}. In this paper, we propose a new method to challenge~this~issue.}
\vspace{-15pt} 
\label{fig:idea}
\end{figure} 

The state-of-the-art domain adaptation methods align the two domains by focusing on either minimizing a divergence metric or confusing a domain discriminator to learn domain-invariant features. For instance, deep CORAL~\cite{sun2016deep} minimizes the covariance between the source features and the target features. JAN~\cite{long2017deep} minimizes the widely used Maximum Mean Discrepancy (MMD) metric~\cite{gretton2012kernel} on the fully connected layers of deep networks to align the two domains. The metric-based methods have been studied for many years. However, there is no much progress on the metric itself. Almost all of the existing approaches work on how to leverage the very few off-the-shelf metrics, e.g., MMD, KL-divergence, and H-divergence~\cite{ben2010theory}, more effectively. On the other hand, adversarial methods such as ADDA~\cite{tzeng2017adversarial} and CoGAN~\cite{liu2016coupled} learn domain-invariant features by adversarially fooling the domain discriminator as illustrated in Fig.~\ref{fig:idea}. Recently, several works~\cite{arora2017generalization,long2018conditional} claim that there is no guarantee~that~the~two domains will be aligned even if the domain discriminator is fully confused, which is caused by the equilibrium challenge~\cite{arora2017generalization,long2018conditional} in adversarial learning. As a result, it is highly expected if there is another way for adaptation which does not only employ the conventional metrics and somewhat can alleviate the equilibrium challenge in~adversarial~learning.

To alleviate the equilibrium challenge in adversarial domain adaptation, a nature idea is that if we can find a divergence metric and optimize it during the adversarial learning. As a result, we can not only confuse the domain discriminator to learn domain-invariant features but also guarantee that the statistical divergence between the two domains is minimized. However, the existing divergence metrics have specific limitations. For instance, the covariance is too simple and optimizing it can only guarantee the second-order statistics rather than the real distributions being aligned; The KL-divergence has been verified that it cannot be generalized to the real distribution when optimizing on the observed mini-batch distributions~\cite{arora2017generalization}; Although MMD~\cite{gretton2012kernel} has been widely used as a distance metric for domain adaptation, it is still unclear how to explicitly minimize the marginal and conditional MMD in deep~neural~networks~\cite{long2017deep}. At the same time, the calculation of MMD generally costs quadratic time, which is not efficient for~large-scale~datasets.

\begin{figure}[t]
\begin{center}
\includegraphics[width=0.8\linewidth]{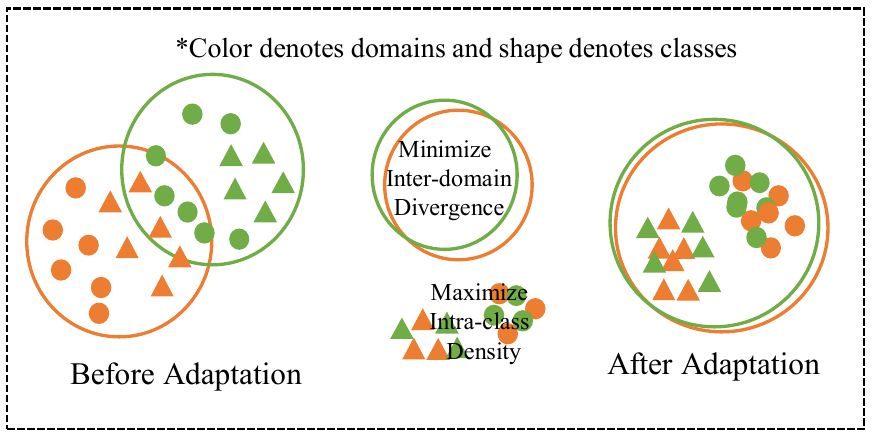}
\end{center}
\vspace{-10pt}
\caption{\small The illustration of our proposed distance loss MDD (Maximum Density Divergence). MDD has two motivations: minimizing the inter-domain divergence and maximizing the intra-class density. }
\label{fig:mdd}
\vspace{-15pt}
\end{figure}

In this paper, on one hand, we propose a new learning loss named Maximum Density Divergence (MDD) which can be used to minimize the domain distribution gaps. The main motivation of MDD is illustrated in Fig.~\ref{fig:mdd}. Specifically, MDD jointly minimizes the inter-domain divergence and maximizes the intra-domain density. Different from the widely-used MMD, we present a practical variant of MDD which can be smoothly and efficiently incorporated into deep domain adaptation architectures. It is worth noting that our MDD not only considers the marginal distribution between domains by minimizing the inter-domain divergence but also optimizes the conditional distribution between domains by introducing class information.

On the other hand, we find that existing adversarial domain adaptation methods~\cite{long2018conditional,hoffman2018cycada,tzeng2017adversarial} suffer the plight of ``there is no guarantee that the two domains are well aligned even if the domain discriminator is fully confused'' because these approaches devote themselves to optimizing the adversarial loss. Since the adversarial loss is highly related with the equilibrium challenge~\cite{arora2017generalization} which is inherent in adversarial training, optimizing only adversarial loss cannot break through the plight. As stated in~\cite{arora2017generalization}, just as a zero gradient is a necessary condition for standard optimization to halt, the corresponding necessary condition in a two-player game is an equilibrium. Conceivably some of the instability often observed while training GANs could just arise due to lack of equilibrium. When this issue comes into domain adaptation, we care about not only the training stability of the model but also the distribution divergence between the two domains. To address this, a reasonable idea is to optimize an additional loss function which can explicitly minimize the distribution shifts. Fortunately, our proposed MDD is a solid choice. We prove that the MDD between two distributions is zero if the two are equivalent. Furthermore, we propose a practical variant of MDD loss which is computationally economic. In summary, the main contributions of this paper can be outlined as:

1) We present a novel loss MDD for unsupervised domain adaptation. MDD jointly minimizes the inter-domain divergence and maximizes the intra-domain density. It can be seamlessly incorporated into deep domain adaptation networks and optimized by stochastic gradient descent. 

2) We argue that the equilibrium challenge issue in adversarial domain adaptation can be alleviated by optimizing an additional loss which quantifies the distribution gaps. In this paper, we deploy the proposed MDD loss. The loss can also be used as a regularization term in adversarial domain adaptation networks to improve the performance.

3) We propose a novel method named Adversarial Tight Match (ATM) for unsupervised domain adaptation. The proposed ATM leverages the MDD in adversarial domain adaptation networks. Experiments on four benchmarks verify that our method can significantly outperform several previous state-of-the-arts. The result on SVHN$\rightarrow$MNIST, for instance, is improved from~89.2\%~to~96.1\%.

The rest of this paper is organized as follows. Section~2 presents a brief review on related work and highlights the merits of our method by comparing with existing ones. Section~3 details the proposed method from MDD to ATM. Section~4 reports the experiments on various datasets which consist of both classical ones and large-scale ones. At last, we conclude this paper in Section~5.

\begin{figure*}[!ht]
\begin{center}
\includegraphics[width=0.71\linewidth]{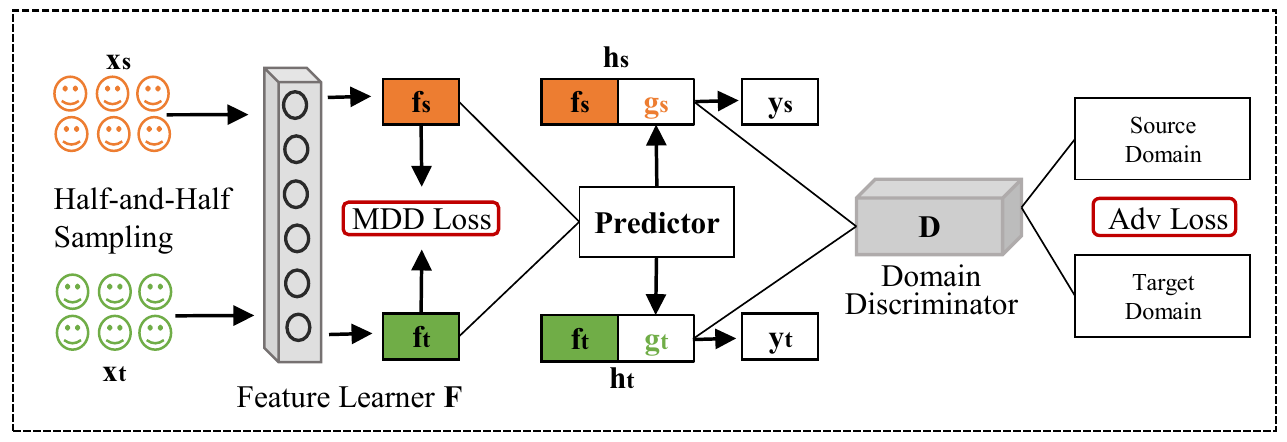}
\end{center}
\vspace{-10pt}
\caption{The idea illustration of the proposed adversarial tight match (ATM) for domain adaptation. Our method simultaneously optimizes the MDD loss and the adversarial loss. As a result, it can not only confuse the domain discriminator but also guarantee that the two data distributions are well aligned. The feature learner $F$ is a deep neural network, e.g., LeNet for the digits recognition and ResNet-50 for the object recognition in this paper. The predictor is a softmax classifier which has two purposes: generating the classification condition $\mathrm{p}$ and predicting the pseudo labels $y_t$ for target samples.}
\label{fig:atm}
\vspace{-10pt}
\end{figure*}

\section{Related Work}
\subsection{Metric Learning for Domain Adaptation}
Domain adaptation~\cite{ganin2016domain,li2018heterogeneous,ding2018semi,jing2020adaptive} addresses the problem of knowledge transfer between two domains which have distinctive data distributions. Therefore, a practical way for domain adaptation is to define a distance metric which can measure the distribution discrepancy between domains. Then, domain adaptation can be formulated as a distance minimization problem~\cite{long2017deep,long2015learning,sun2016deep,ben2010theory}. 

In existing works, four distance metrics have been explored: 1) MMD~\cite{long2017deep,long2015learning,gretton2012kernel}, 2) Covariance~\cite{sun2016deep}, 3) H-divergence~\cite{ben2010theory} and 4) Kullback–Leibler (KL) divergence~\cite{zhuang2015supervised}. Specifically, MMD measures the difference between the mean function values on two samples and proves that the population MMD is zero if and only if the two distributions are equivalent. MMD has been widely used in conventional domain adaptation approaches~\cite{li2018transfer,li2019locality}. For instance, Li et al.~\cite{li2018transfer} propose a unified framework for domain adaptation based on MMD minimization and landmark selection. When it comes to deep models, however, it is still unclear how to explicitly optimize conditional MMD. The representation of covariance-based methods is Deep CORAL~\cite{sun2016deep}, which is simple yet effective in several cases. However, covariance-based methods are not substantially optimal for domain adaptation problems where the distribution divergences are large. H-divergence was introduced as a classifier-induced divergence. The most important merit of H-divergence is that it can be estimated from finite samples~\cite{ganin2016domain}. In recent work, H-divergence is generally used as a tool to analyze the generalization bound of proposed methods rather than directly optimizing it. 
In this paper, we propose a new divergence MDD, which is mainly compared with MMD. It is worth noting that our method jointly optimizes the inter-domain divergence and the intra-class density. The MDD loss, to some extent, is related with jointly optimizing the marginal MMD and the conditional MMD. At last, the proposed MDD is related with the energy distance~\cite{szekely2003statistics}. However, our motivation is significantly different from the energy distance. In addition, the optimizations of inter-domain divergence and the intra-class density are in direct contradiction in the formulation of energy distance.

\subsection{Adversarial Domain Adaptation}
The idea of adversarial domain adaptation~\cite{tzeng2017adversarial,long2018conditional,li2019cycle} is different from metric learning which explicitly optimizes a similarity function. Instead, the idea of adversarial domain adaptation is somewhat heuristic. It trains a domain discriminator along with a feature learning network via an adversarial manner. The feature learning network learns domain-invariant features and tries to fool the domain discriminator. Once the domain discriminator is fully confused, i.e., unable to tell whether an instance is from the source domain or the target domain, the adversarial domain adaptation assumes that the distribution gaps between the two domains are well aligned. For instance, Tzeng et al.~\cite{tzeng2017adversarial} propose adversarial discriminative domain adaptation (ADDA) which combines discriminative modeling, untied weight sharing and a GAN~\cite{goodfellow2014generative} loss in a general framework. Li et al.~\cite{li2019cycle} further introduce cycle-consistent loss into the framework and~achieve~state-of-the-art~results.

Recently, Long et al.~\cite{long2018conditional} point out that a vanilla adversarial domain adaptation framework cannot guarantee that the two domains are well aligned even if the domain discriminator has been fully confused. Such a claim is based on the equilibrium challenge of adversarial learning~\cite{arora2017generalization}. The challenge in GANs is formally defined in~\cite{arora2017generalization}.

Based on the definition, Arora et al. propose to train the GANs with multiple discriminators and multiple generators. However, the paradigm needs to train multiple models. In the domain adaptation community, we generally compare the performance with single models. To address this issue, Long et al.~\cite{long2018conditional} propose conditional domain adversarial network (CDAN) by sharing the spirit of the conditional GANs~\cite{mirza2014conditional}. Recently, there are several studies~\cite{shu2018dirt,zhao2019learning} performed on adversarial training which aim to address the divergence in the conditional distribution. For instance, Zhao et al.~\cite{zhao2019learning} consider the conditional shift and prove an information-theoretic lower bound on the joint error of any domain adaptation method that attempts to learn invariant representations. Shu et al.~\cite{shu2018dirt} introduce cluster assumption, i.e., decision boundaries should not cross high-density data regions, into adversarial domain adaptation to address the existing issues. Xie et al.~\cite{xie2018learning} learn semantic representations by aligning labeled source centroid and pseudo-labeled target centroid. However, our formulation is significantly different from~existing~ones.

\section{The Proposed Method}
In this section, we first introduce our novel MDD. Then, we prove it is a lower bound of symmetric KL-divergence. At last, we present a practical variant of the MDD loss and show how to incorporate it into adversarial domain adaptation framework and propose our new method ATM. A global glance at the proposed ATM is illustrated in Fig.~\ref{fig:atm}.

\renewcommand\arraystretch{1.5}
\subsection{Maximum Density Divergence}
\begin{figure}[t]
\begin{center}
\includegraphics[width=0.73\linewidth]{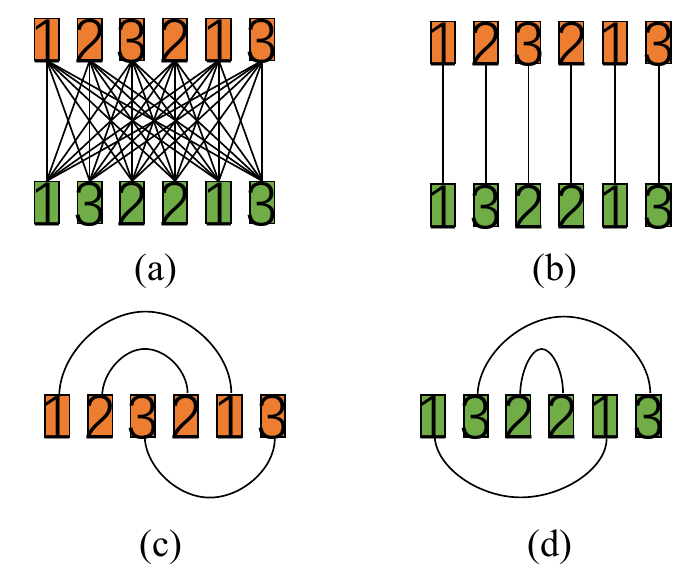}
\end{center}
\vspace{-15pt}
\caption{\small Idea illustration for Eq.~\eqref{eq:bddistance}. Better viewed in color. We use orange for source domain and green for target domain. The number in each box denotes the class information. A line means that the distance between the two samples will be considered into the loss. For clarity, we use $5$ samples per domain in a batch for this example. To calculate the first term in Eq.~\eqref{eq:ddistance}, we need to calculate it in the way shown in (a). In this paper, we simplify it as shown in (b), i.e, the first term in Eq.~\eqref{eq:bddistance}. The second and the third terms are calculated as shown in (c) and (d), respectively. }
\label{fig:eq6}
\vspace{-10pt}
\end{figure}

Suppose we have a source domain $\mathbb{D}_s$ and a target domain $\mathbb{D}_t$ where samples in the two domains are denoted as $X_s$ and $X_t$, respectively. The data probability distributions of $X_s$ and $X_t$ are $P$ and $Q$, respectively, i.e., $X_s \sim P$ and $X_t \sim Q$. In domain adaptation, we have $P\neq Q$. Our goal is to align the data probability distributions of the two domains. If we use $\mathrm{MDD}(P,Q)$ to quantify the distribution gaps of $P$ and $Q$, the target of domain adaptation can be formulated as minimizing $\mathrm{MDD}(P,Q)$. In this paper, we define a novel implementation of $\mathrm{MDD}(P,Q)$ as follows:
\begin{equation}
\label{eq:distance}
  \begin{array}{c}
 \mathrm{MDD}(P,Q)=\mathbb{E}_{X_s \sim P, X_t \sim Q}[\|X_s-X_t\|^2_2] + \\ \mathbb{E}_{X_s,X'_s \sim P}[\|X_s-X'_s\|_2^2]+\mathbb{E}_{X_t,X'_t \sim Q}[\|X_t-X'_t\|_2^2],
  \end{array} 
\end{equation}
where $X'_s$ is an independent and identically distributed (iid) copy of $X_s$, and $X'_t$ is an iid copy of $X_t$. From Eq.~\eqref{eq:distance}, we can see that the proposed MDD consists of three parts. Intuitively, the first term minimizes inter-domain divergence between $P$ and $Q$. The left two terms maximize the intra-domain density of $P$ and $Q$, respectively. As a result, we can not only align the two domains but also make the two domain themselves more compact, which is why we name the proposed metric as maximum density divergence. In this paper, we deploy the squared Euclidean norm distance as shown in Eq.~\eqref{eq:distance}. In practice, the distance can be generalized to different variations according to different target tasks. Specifically, we have the generalized form of MDD as:
\begin{equation}
  \begin{array}{c}
 \mathbb{E}_{X_s \sim P, X_t \sim Q} |X_s-X_t|_\ell+ \mathbb{E}_{X_s,X'_s \sim P}|X_s-X'_s|_\ell \\+ \mathbb{E}_{X_t,X'_t \sim Q}|X_t-X'_t|_\ell,
  \end{array} 
\end{equation}
where $|\cdot|_\ell$ indicates different norms. In domain adaptation, the probability $P$ and $Q$ are defined on point sets rather than cumulative distribution functions (CDF). Suppose we have $n_s$ samples $\{x_{s,1},x_{s,2},\cdots,x_{s,n_s}\}$ in the source domain and $n_t$ samples $\{x_{t,1},x_{t,2},\cdots,x_{t,n_t}\}$ in the target domain, and their labels are denoted as $y_s$ and $y_t$, respectively. Then, we can rewrite Eq.~\eqref{eq:distance} as:
\begin{equation}
\label{eq:ddistance}
  \begin{array}{c}
 \!\!\!\!\frac{1}{n_s n_t}\sum_{i,j}^{n_s,n_t}\|x_{s,i} \!-\! x_{t,j}\|^2_2 \!+\! \frac{1}{n_s n_s}\sum_{i,j}^{n_s,n_s}\|x_{s,i} \!-\! x'_{s,j}\|_2^2 \\ + \frac{1}{n_t n_t}\sum_{i,j}^{n_t,n_t}\|x_{t,i} \!-\! x'_{t,j}\|_2^2,
  \end{array} 
\end{equation}
where $i$ and $j$ denote the index of samples. In domain adaptation tasks, there are several challenges when calculating Eq.~\eqref{eq:ddistance}. The first challenge is that Eq.~\eqref{eq:ddistance} considers all the pair-wise distances between samples. However, deep networks are generally trained batch by batch. We can only obtain a fraction of whole dataset in each batch rather than all the data at the same time. The second challenge is how to explicitly sample $x'_s$ and $x'_t$. To make the proposed MDD practical in real-world domain adaptation tasks, we propose two solutions to address these two challenges. The first solution is that we only calculate the pair-wise distance at the relative position rather than all the distances in the first term. The second solution is that we sample $x'_s$ and $x'_t$ as the samples which have the same label with $x_s$ and $x_t$ within a batch. As a result, we tailor Eq.~\eqref{eq:ddistance} as the following form:
\begin{equation}
\label{eq:bddistance}
  \begin{array}{c}
 \frac{1}{n_b}\sum_{i}^{n_b}\|x_{s,i} - x_{t,i}\|^2_2 +\frac{1}{m_s}\sum_{y_{s,i}=y'_{s,j}}\|x_{s,i}- x'_{s,j}\|_2^2 \\ + \frac{1}{m_t}\sum_{y_{t,i}=y'_{t,j}}\|x_{t,i} - x'_{t,j}\|_2^2,
  \end{array} 
\end{equation}  
where $n_b$ is equal to the half of the {\it batch size}, $y_{s,i}=y'_{s,j}$ indicates that $x_{s,i}$ and $x'_{s,i}$ have the same label. It is worth noting that: 1) we deploy half-and-half sampling for the source domain and target domain in a given batch and only calculate the pair-wise distance at relative position for $\|x_{s,i} \!-\! x_{t,j}\|^2_2$ to speed up the calculation, so that the coefficient of the first term is $1/n_b$ ; 2) the number of samples in a batch which satisfies $y_{s,i}=y'_{s,j}$ is uncertain before training. Thus, we use $m_s$ and $m_t$ to represent the appropriate number in the source domain and the target domain, respectively. It is worth noting that $m_s$ and $m_t$ can be determined in each batch by checking the labels of each instance in the current batch. In fact, we often do not need to calculate $m_s$ and $m_t$ in practice. Almost all of the programming platforms provide the {\it average} operation.  For a better understanding, we illustrate the idea of Eq.~\eqref{eq:bddistance} in Fig.~\ref{fig:eq6}.

In practice, we sample $n_b$ source samples and $n_b$ target samples in one batch. If the total numbers of the source samples and target samples are mismatched, we would sample them in a round-robin way to guarantee the sampling. Since the deep networks are trained in an iterative manner, we only calculate the pair-wise distance at relevant position for $\|x_{s,i} \!-\! x_{t,j}\|^2_2$ to simplify the calculation. For instance, we only calculate $\|x_{s,1} \!-\! x_{t,1}\|^2_2$, $\|x_{s,2} \!-\! x_{t,2}\|^2_2$, $\cdots$, and $\|x_{s,n_b} \!-\! x_{t,n_b}\|^2_2$. With the fact that samples are randomly permuted, the simplified way does not affect the final performance. In the experiments, we will show that such a way can also lead to convergence within feasible iterations. It is also worth noting that we cannot access $y_t$ in unsupervised domain adaptation. In order to calculate the third term, we use pseudo labels which are jointly predicted by our domain adaptation network. In the experiments, we will discuss the accuracy of pseudo labeling.

Now, we investigate the theoretical properties of our proposed MDD loss function. We prove that Eq.~\eqref{eq:distance} is a lower bound of symmetric KL-divergence. Furthermore, we also report the low-bound of our MDD. \\

\noindent {\bf Lemma 1.} {\it On a finite probability space, the proposed MDD in Eq.~\eqref{eq:distance} is a lower bound of symmetric KL-divergence.} 

\noindent {\bf Lemma 2.} {\it On a finite probability space, the proposed MDD in Eq.~\eqref{eq:distance} is bounded by $4\delta^2(P,Q)$, where $\delta(P,Q)$ is the total variation distance between $P$ and $Q$.} 

\begin{proof}
\begin{small}
For a finite-dimensional vector $x$, we have $\|x\|_2\le\|x\|_1$. Therefore,
% \begin{equation}
% \begin{array}{l}
$$\mathrm{MDD}(P,Q)=\mathbb{E}_{X_s \!\sim\! P, X_t \!\sim\! Q}[\|X_s\!-\!X_t\|^2_2]\! +\!\mathbb{E}_{X_s,X'_s \!\sim\! P}[\|X_s\!-\!X'_s\|_2^2]\! $$$$\\ ~~~~~~~~~~~~~~~~~~~ + \!\mathbb{E}_{X_t,X'_t \!\sim\! Q}[\|X_t\!-\!X'_t\|_2^2] $$$$\\ ~~~~~~~~~~~~~~~~~~\le \mathbb{E}_{X_s \!\sim\! P, X_t \!\sim\! Q}[\|X_s\!-\!X_t\|^2_1]\!+\!\mathbb{E}_{X_s,X'_s \!\sim\! P}[\|X_s\!-\!X'_s\|_1^2]\! $$$$\\ ~~~~~~~~~~~~~~~~~~~+\!\mathbb{E}_{X_t,X'_t \!\sim\! Q}[\|X_t\!-\!X'_t\|_1^2]$$
% \end{array}
% \end{equation}

Considering that both $X_s$ and $X_t$ are countable, the total variation distance between $P$ and $Q$ is $\delta(P,Q)=\frac{1}{2}\|P-Q\|_1$~\cite{levin2017markov}, we have:
$$\mathrm{MDD}(P,Q)\le 4\delta^2(P,Q) = 2\delta^2(P,Q)+2\delta^2(Q,P).$$
Further, by Pinsker's Inequality $\delta(P,Q)\le \sqrt{\frac{1}{2}D_{KL}(P||Q)}$ and $\delta(Q,P)\le \sqrt{\frac{1}{2}D_{KL}(Q||P)},$ we have:
$$\mathrm{MDD}(P,Q)\le D_{KL}(P,Q)+D_{KL}(Q,P),$$
where the right side is Jeffreys' J-divergence, a symmetric Kullback-Leibler divergence. At the same time, we can observe that on a finite probability space, the divergence in Eq.~\eqref{eq:distance} is bounded by $4\delta^2(P,Q)$.
\end{small}
\end{proof}

\noindent {\bf Lemma 3.} {\it On a finite probability space, the divergence $\mathrm{MDD}$$(P,Q)=0$ if $P=Q$.} 
\begin{proof}
\vspace{-5pt}
\begin{small}
The KL-divergence is always non-negative and it is 0 only if $P=Q$. In addition, MDD is always non-negative. Thus, with the conclusion of Lemma 1, we have $\mathrm{MDD}(P,Q)= 0$ if $P = Q$. 
\end{small}
\end{proof}

 In fact, the conclusion of Lemma~3 can also be observed by directly investigating Eq.~\eqref{eq:distance}. It is worth noting that we introduce the MDD on the original feature space without loss of generality. In real-world domain adaptation tasks, we may need to align the two domains on the feature space rather than the original space. Therefore, in our proposed domain adaptation method ATM, we calculate the MDD on the learned domain-invariant features $\mathrm{f}_i=F(x_i)$, where $F$ is the feature representation network. The overall idea of our proposed ATM is illustrated in Fig.~\ref{fig:atm}. In practice, we calculate the MDD loss by the following equation: 
\begin{equation}
\label{eq:mddloss}
  \begin{array}{l}
\!\!\! \mathcal{L}_{mdd}=\frac{1}{n_b}\!\sum_{i}^{n_b}\!\|\mathrm{f}_{s,i} \!-\! \mathrm{f}_{t,i}\|^2_2 \!+\!\frac{1}{m_s}\!\sum_{y_{s,i}=y'_{s,j}}\!\!\|\mathrm{f}_{s,i}\! \!-\! \mathrm{f}'_{s,j}\|_2^2 \! \\~~~~~~~~~~~~~+ \frac{1}{m_t}\!\sum_{y_{t,i}=y'_{t,j}}\!\!\|\mathrm{f}_{t,i} \!-\! \mathrm{f}'_{t,j}\|_2^2,
  \end{array}
\end{equation}  
where $n_b$ is the half of the batch size, $m_s$ and $m_t$ can be dynamically calculated in each batch.

It is worth noting that Eq.~\eqref{eq:mddloss} considers both the inter-domain divergence and the intra-domain density. Thus, it takes care of both marginal distribution and conditional distribution. Since the third term in Eq.~\eqref{eq:mddloss} involves pseudo labeling, one may worry about the accuracy of the pseudo labels. In fact, pseudo labeling is an effective strategy in unsupervised learning~\cite{xie2018learning}. The pseudo labels are not fixed after initialization. They are dynamically updated in each training epoch. In the experiments, we will show that the accuracy of the pseudo labeling is steadily increased with the increase of training iterations until convergence.

\begin{figure*}[t]
\begin{center}
\includegraphics[width=0.73\linewidth]{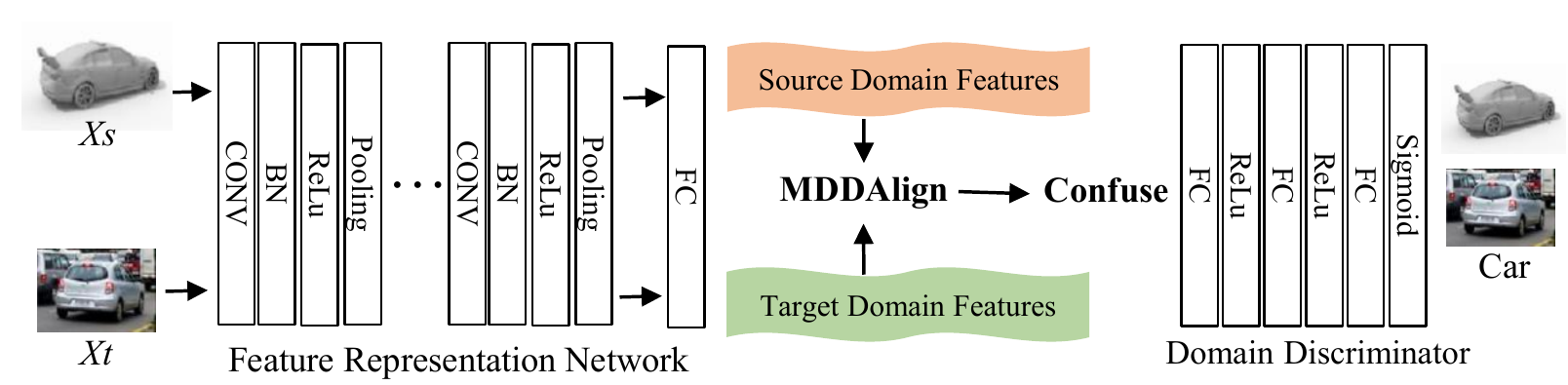}
\end{center}
\vspace{-10pt}
\caption{An illustration of the implementation of the proposed method. In this paper, we mainly use ResNet-50 as the~feature representation network (exceptions are stated in the context, e.g., LeNet is used for digits recognition to avoid over-fitting). The domain discriminator is implemented by three FC layers. The MDD alignment is implemented according to Eq.~\eqref{eq:mddloss}.}
\label{fig:implement}
\vspace{-10pt}
\end{figure*}

\subsection{Cross-Domain Adversarial Tight Match}
The idea of adversarial domain adaptation is similar to generative adversarial networks (GANs)~\cite{goodfellow2014generative}. In adversarial domain adaptation~\cite{tzeng2017adversarial,long2018conditional,hoffman2018cycada}, there are also a generator and a discriminator. The difference is that the generator does not synthesize fake instances from noise. In fact, it is a feature representation network which learns domain-invariant features from both the source domain and the target domain. At the same time, the discriminator does not distinguish from real to~fake.~It~tells~source domain features from target domain features. The feature representation network and the domain discriminator are trained in an adversarial manner. Once the domain discriminator cannot distinguish whether a learned feature belongs to the source domain or the target domain, it is considered that the learned representations are domain-invariant. In this paper, we use $F$ to denote the feature learner and $D$ to denote the domain discriminator. The implementations of $F$ and $D$ are shown in Fig.~\ref{fig:implement}. Formally, we can formulate the adversarial domain adaptation networks as follows:
\begin{equation}
\label{eq:cdan}
  \begin{array}{l}
 %\min\limits_G \max\limits_D \mathbb{E}[log D(x,a)]+ 
 \min\limits_{F}\max\limits_{D} \mathcal{L}_{adv}\!=\! -\mathbb{E}[\sum_{c=1}^{C}\mathbbm{1}_{[y_s=c]}\mathrm{log} \sigma(F(x_s))]  \\~~~~~~~~~~~~~~~~~~~~ + \lambda (\mathbb{E}[\mathrm{log}D(\mathrm{h}_s)]+\mathbb{E}[\mathrm{log}(1-D(\mathrm{h}_t))]),
\end{array} 
\end{equation} 
where $\lambda>0$ is a balancing parameter. In this paper, we follow CDAN~\cite{long2018conditional} and fix $\lambda=1$ for fair comparisons. The first term in Eq.~\eqref{eq:cdan} is a supervised cross-entropy loss on the source domain, in which $\mathbbm{1}_{[\cdot]}$ is an indicator, $\sigma$ is the softmax and $C$ is the possible categories. The second term is a conditional loss which is very similar to conditional GAN~\cite{mirza2014conditional}. It is worth noting that $\mathrm{h=\Pi(f,p)}$ is the joint variable of the domain specific features $\mathrm{f}$ and its corresponding classification predictions $\mathrm{p}$. Specifically, the feature $\mathrm{f}$ is learned by the feature learner $F$, i.e., $\mathrm{f}=F(x)$. The classification predictions are calculated by a softmax classifier. $\Pi(\cdot)$ is the conditioning strategy. The idea of conditioning with classifier predictions is inspired by previous work CDAN. One can refer to~\cite{long2018conditional} for more details. In this paper, we deploy the entropy condition since it generally performs better than linear condition. In the experiments, we also report the results on CDAN with the entropy condition for fair compassions.

Optimizing Eq.~\eqref{eq:cdan} has shown promising performance in domain adaptation. However, it still suffers from the equilibrium challenge, i.e., there is no guarantee that the two domains are well aligned even the domain discriminator is fully confused. In this paper, we propose to explicitly align the two distributions when confusing the domain discriminator. In other words, we simultaneously minimize the distance metric and confuse the domain discriminator. Since the two sides have the same purpose, simultaneously optimizing them can reinforce each other. With an appropriate divergence measure, we can make sure that the distributions will be well aligned when the domain discriminator is confused. Therefore, we introduce our proposed Maximum Density Divergence (MDD) into the adversarial domain adaptation framework. As a result, the overall objective function of our method ATM can be written as:
\begin{equation}
\label{eq:overall}
  \begin{array}{l}
 \min\limits_{F}\max\limits_{D} \mathcal{L}_{adv} + \alpha \mathcal{L}_{mdd}  ,
\end{array} 
\end{equation} 
where $\alpha>0$ is a balancing parameter, $\mathcal{L}_{mdd}$ is the MDD loss which can be calculated in each batch by Eq.~\eqref{eq:mddloss}.~The~main ideas of ATM are illustrated in Fig.~\ref{fig:atm}~for~better~understandings. For clarity, we sketch the main steps of our method in {\it Algorithm~1}. We also present the implementation in Fig.~\ref{fig:implement}.

At last, it is worth noting that entropy minimization is widely used in domain adaptation to learn discriminative representation. In this paper, on the contrary, we leverage entropy maximization to confuse the domain discriminator.

\renewcommand\arraystretch{1}
\begin{table}[t!p]
\begin{small}
    \begin{center}
    % \caption{Algorithm1} \label{tab:cap}
    \begin{tabular}{p{8.5cm}<{\raggedright}}
      \toprule
        { {\bf Algorithm~1: Cross-Domain Adversarial Tight Match}\;} \\
      \midrule
        {\bf Input:} Source and target domain data $X_s$ and $X_t$; labels for source domain data $y_{\rm s}$; parameters $\alpha$ and $\lambda=1$. \\
        {\bf Output:} Predicted labels $y_t$ for target domain unlabeled data. \\
        {\bf begin} \\           
        ~~~~{\bf while} not converge and epoch < max\_epoch {\bf do}\\
            ~~~~~~~~1. Randomly Sample $n_b$ labeled source domain instances and $n_b$ unlabeled target domain instances.\\
            ~~~~~~~~2. Using the feature representation network $F$ to learn features for the samples.\\
            ~~~~~~~~3. Train the classifier and get the pseudo labels of $X_t$.\\
            ~~~~~~~~4. Calculate the learning loss in Eq.~\ref{eq:overall}.\\
            ~~~~~~~~5. Update the networks $F$ and $D$ by mini-batch SGD.\\
        ~~~~{\bf end while}\\
        {\bf end} \\
      \hline
    \end{tabular}
    \end{center}
    \vspace{-10pt}
    \end{small}
    \end{table}

\subsection{Generalization Bound Analysis}

For a better understanding of our work, we report a brief theoretical analysis based on the widely used Ben-David theory of domain adaptation~\cite{ben2010theory}. Formally, the Ben-David theory is defined as follows:

\begin{theorem}
Let $\mathcal{H}$ be a hypothesis space. If we denote the generalization errors of a function $f\in \mathcal{H}$ on the target domain $X_t$ and the source domain ${X}_s$ as $\epsilon_t$ and $\epsilon_s$, respectively. Then, for any function $f\in \mathcal{H}$,
\begin{equation}
  \begin{array}{c}
 \epsilon_t(f)\le \epsilon_s(f)+d_{\mathcal{H}\Delta\mathcal{H}}(X_s,X_t)+\epsilon^*,
  \end{array} 
\end{equation}
where $d_{\mathcal{H}\Delta\mathcal{H}}(X_s, X_t)$ is $\mathcal{H}\Delta\mathcal{H}$-distance which measures the discrepancy between the two domains:
\begin{equation}
  \begin{array}{c}
 d_{\mathcal{H}\Delta\mathcal{H}}(X_s, X_t)=\sup\limits_{f,f'\in \mathcal{H}}\big|\mathbb{E}_{x\sim X_s}[f(x)\neq f'(x)]\\~~~~~~~~~~~~~~~~~~~~~~~~~~~~~~~-\mathbb{E}_{x\sim X_t}[f(x)\neq f'(x)]\big|,
  \end{array} 
\end{equation}
and $\epsilon^*$ is the shared error of an ideal joint hypothesis $f^*$,
\begin{equation}
  \begin{array}{l}
  f^*=\arg\min_{f\in\mathcal{H}}~\epsilon_s(f)+\epsilon_t(f), \\
  \epsilon^* = \epsilon_s(f^*)+\epsilon_t(f^*).
  \end{array} 
\end{equation}
\end{theorem}

From {\it Theorem~1}, we can see that the generalization error on the target domain is bounded by three aspects: (1) the expected error on the source domain; (2) the domain discrepancy between the source domain and the target domain and (3) the shared error of the ideal joint hypothesis. Now, we analyze why our model is able to minimize the generalization error on the target domain.

For the first term $\epsilon_s(f)$, our model minimizes $\mathbb{E}[\mathrm{log} P(y_s|x_s)])$, which explicitly reduces the classification error on the source domain. Regarding $d_{\mathcal{H}\Delta\mathcal{H}}(X_s, X_t)$, from the analysis in~\cite{ganin2014unsupervised}, we have
\begin{equation}
  \begin{array}{c}
 d_{\mathcal{H}\Delta\mathcal{H}}(X_s, X_t)\le 2\sup\limits_{D\in \mathcal{H}_D}\big|D(\mathrm{h})-1\big|,
  \end{array} 
\end{equation}
where $D$ is the domain discriminator, $D(\mathrm{h})$ is maximized by the optimal $D$. Notice that this conclusion is based on the assumption that we have a freedom to pick $\mathcal{H}_D$. Fortunately, a multilayer neural network is able to fit any functions. As long as the feature representation network learns the domain-invariant features, the $d_{\mathcal{H}\Delta\mathcal{H}}(X_s, X_t)$ tends to decrease during the adversarial training. At the same time, our MDD actually explicitly minimizes $d_{\mathcal{H}\Delta\mathcal{H}}(X_s, X_t)$ during the optimization. At last, the last term $\epsilon^*$ does not depend on particular $f$.

For the equilibrium issue, we experimentally verify it in the next section. We will show that our method is better towards the convergence. Here, we give a brief explanation based on the theories in~\cite{arora2017generalization}. In~\cite{arora2017generalization}, challenging the equilibrium is that the generator ``wins'' meaning the discriminator cannot do better than random guessing. The solution given by Arora et al.~\cite{arora2017generalization} is to train a mixture of generators and discriminators. However, this strategy needs huge GPU memory. It is also not fair to compare with other single model methods. In this paper, we introduce the MDD loss as a regularization for the adversarial training. The regularization of course affects the parameter updating of the adversarial networks. In fact, the very motivation behind training mix-GANs in~\cite{arora2017generalization} is to search more parameters with different generators and discriminators. With the regularization of our MDD, the generator and discriminator are prone to maintain the parameters which are able to learn domain-invariant representations. The gradients are updated towards the direction where the two domains can be well aligned.

\section{Experiments}
In this section, we evaluate the proposed method on four widely used datasets which consist of both standard and large-scale datasets. State-of-the-art methods reported in recently years are compared. 

\subsection{Data Preparation}
{\bf MNIST}, {\bf USPS} and Street View House Numbers ({\bf SVHN})~\cite{netzer2011reading}, are three widely used handwritten digits dataset. MNIST (M) consists of 60,000 training samples and 10,000 test samples. USPS (U) is comprised of 7,291 training samples and 2,007 test samples. SVHN (S) contains over 600,000 labeled digits cropped from street view images. 

{\bf Office-31}~\cite{saenko2010adapting} consists of 3 subsets, i.e., {Amazon} (A), {Webcam} (W) and {DSLR} (D). Specifically, the images in amazon are downloaded from amazon.com. Webcam and DSLR contain images captured by a web camera and an SLR camera, respectively. In total, there are 4,652 samples~from~31~categories. 

{\bf ImageCLEF-DA}\footnote{http://imageclef.org/2014/adaptation} consists of 12 common classes shared by Caltech-256 (C), ImageNet ILSVRC 2012 (I) and Pascal VOC 2012 (P). Specifically, the 12 classes are aeroplane, bike, bird, boat, bottle, bus, car, dog, horse, monitor, 
motorbike and people.

{\bf Office-Home}~\cite{venkateswara2017Deep} consists of images from 4 different domains: Artistic images (A), Clip Art (C), Product images (P) and Real-World images (R). For each domain, the dataset contains images of 65 object categories found typically in office and home settings.

\subsection{Experiment Protocols}

As shown in Fig.~\ref{fig:atm}, our model mainly consists of three components, i.e., the feature learner $F$, the predictor and the domain discriminator $D$, and two training losses, i.e, the MDD loss and the adversarial loss. The feature learner $F$ is implemented by convolutional neural nets. For instance, we mainly deploy ResNet-50 as the backbone network of $F$ except for digits recognition where we follow the same settings in CyCADA~\cite{hoffman2018cycada}. Specifically, the $F$ implemented in digits recognition is a variant of the classical LeNet. The predictor is implemented by a standard softmax classifier. The domain discriminator $D$ is implemented by FC-ReLU-FC-ReLU-FC-Sigmoid. We illustrate the main implementations in Fig.~\ref{fig:implement}. For the two training losses, the MDD loss is calculated by Eq.~\eqref{eq:mddloss} and the adversarial loss is calculated by Eq.~\eqref{eq:cdan}, more details of the adversarial loss can also be found in~\cite{long2018conditional}.

We implement our method by PyTorch and optimize our model by mini-batch stochastic gradient descent (SGD) with a weight decay of $5\times10^{-4}$ and momentum of $0.9$. For digits recognition on MNIST, USPS and SVHN, we use the same setting in CyCADA~\cite{hoffman2018cycada}. Specifically, $60000$, $7291$ and $73257$ images from MNIST, USPS and SVHN, respectively, are used for training. The basic network structure for digits recognition is similar to LeNet. We set the batch size to $224$ and the learning rate as $10^{-3}$. For object recognition on Office-31 and ImageCLEF-DA, we follow the same settings in CDAN~\cite{long2018conditional}. The basic network is ResNet-50~\cite{he2016deep} which is pre-trained on ImageNet~\cite{ILSVRC15}. The batch size is 32 and the learning rate is adjusted by the same strategy reported in CDAN. For the experiments on Office-Home, we also deploy ResNet-50 as the base architecture and follow the same settings in~\cite{long2018conditional}. The main steps of our method are reported in {\it Algorithm~1}.

\renewcommand\arraystretch{1}
\begin{table}[t]
\centering
\footnotesize
\vspace{5pt}
\caption{Accuracy (\%) of digits recognition. The best results (single models) are highlighted by bold numbers. The line ``boost $\uparrow$'' shows the accuracy improvements over CDAN. The baseline results on M$\rightarrow$S are cited from SBADA-GAN~\cite{russo2018source}. $^*$Please notice that both SBADA-GAN and Self-Ensembling deploy the ensemble trick. The other methods only use a single model. }
\label{tab:digits}
\vspace{-5pt}
\begin{tabular}{lcccc}
\toprule
%\tabucline[0.75pt]{-----}
Method & M$\rightarrow$U & U$\rightarrow$M & S $\rightarrow$M & M $\rightarrow$S \\
\midrule
Source only & $82.2 $ & $69.6 $ & $67.1 $ & $26.0 $  \\
DANN~\cite{ganin2016domain} & $-$ & $-$ & $73.6$ & $35.7 $  \\
%\hline
ADDA~\cite{tzeng2017adversarial}  & $89.4 $ & $90.1 $ & $76.0 $ & $- $  \\
%\hline
DRCN~\cite{ghifary2016deep} & $91.8 $ & $73.7$ & $82.0 $  & $40.1 $ \\     
%\hline
RevGrad~\cite{ganin2016domain} & $89.1 $ & $89.9 $ & $-$ & $- $  \\
%\hline
CoGAN~\cite{liu2016coupled} & $91.2$ & $89.1$ & $-$ & $- $  \\
%UNIT~\cite{liu2017unsupervised} & $95.9$ & $93.6$ & $90.5$ & $-$ \\
CyCADA~\cite{hoffman2018cycada} & 95.6 & 96.5 & 90.4 & $-$\\
%\hline
%CyCADA~\cite{hoffman2018cycada} & 95.6 & 96.5 & 90.4 & $- $& $-$ \\
CDAN~\cite{long2018conditional} & $95.6$ & $98.0$ & $89.2$& $71.3$  \\ 
\midrule
{ATM [Ours]}  & ${\bf 96.1 }$ & ${\bf 99.0}$ & ${\bf 96.1 }$ & ${\bf 76.6 }$ \\
boost $\uparrow$ & $\uparrow$ 0.5 & $\uparrow$ 1.0 & $\uparrow$ 5.9 & $\uparrow$ 5.3  \\ 
\midrule
{Target Supervised} & 96.3 & 99.2 & 99.2 & 96.7 \\
$^*$SBADA-GAN~\cite{russo2018source} & 97.6 & 95.0 & 76.1 & 61.1 \\
$^*$Self-Ensembling~\cite{french2017self} & 98.3 & 99.5 & 99.2  & 42.0 \\

\bottomrule
\end{tabular}%}
\vspace{-10pt}
\end{table}

In this paper, we deploy the standard unsupervised domain adaptation settings which were widely used in previous works~\cite{long2018conditional,li2018transfer}. Labeled source domain samples $\{X_s,Y_s\}$ and unlabeled target domain samples $X_t$ are used for training. The reported results are the classification accuracy on target samples:

\begin{equation}
  \begin{array}{c}
  accuracy = \frac{|x:~x \in X_t ~\wedge~ \hat{y_t}=y_t|}{|x:~ x \in X_t|},
  \end{array} 
\end{equation}
where ${\hat{y_t}}$ is the predicted label of the target domain generated by our model, and ${y}_t$ is the ground truth label vector. 

The results of the baselines are cited from the original papers, CyCADA and CDAN. It is worth noting that~ the results of CDAN in this paper are all from CDAN+E~in~the original paper since CDAN with entropy~(+E) condition~performs better than CDAN with multi-linear condition. We~do not specify CDAN+E from CDAN for the sake~of~simplicity.

\begin{table*}[ht!p]
\centering
\caption{Domain adaptation results (accuracy~\%) on Office-31. In the header, average~1 is the overall average and average~2 is the average over 4 challenging evaluations except for W$\rightarrow$D and D$\rightarrow$W. ResNet-50 is used for feature learning.}%The best results are highlighted by bold numbers.
\label{tab:office}
\footnotesize
\vspace{-5pt}
\begin{tabular}{lcccccc|cc}
\toprule
Method & A$\rightarrow$D & A$\rightarrow$W & D$\rightarrow$A & D$\rightarrow$W & W$\rightarrow$A & W$\rightarrow$D & Avg~1 & Avg~2\\
\midrule
ResNet~\cite{he2016deep} & $68.9 \pm 0.2$ & $ 68.4 \pm 0.2$ & $62.5 \pm 0.3$ & $96.7 \pm 0.1$ & $60.7 \pm 0.3$ & $99.3 \pm 0.1$ & $76.1$ & $65.1$\\ 
%\hline
TCA~\cite{pan2011domain}   & $74.1 \pm 0.0$ & $72.7 \pm 0.0$ & $61.7 \pm 0.0$ & $96.7 \pm 0.0$ & $60.9 \pm 0.0$ & $99.6 \pm 0.0$ & $77.6 $ & $67.4$ \\
%\hline
%GFK~\cite{gong2012geodesic} & $74.5 \pm 0.0$ & $72.8 \pm 0.0$ & $63.4 \pm 0.0$ & $95.0 \pm 0.0$ & $61.0 \pm 0.0$ & $98.2 \pm 0.0$  & $77.5$ & $67.9$\\  
%\hline
DDC~\cite{tzeng2014deep} & $76.5 \pm 0.3$ & $75.6 \pm 0.2$ & $62.2 \pm 0.4$ & $96.0 \pm 0.2$ & $61.5 \pm 0.5$ & $98.2 \pm 0.1$ & $78.3$ & $69.0$ \\
%\hline
DAN~\cite{long2015learning} & $78.6 \pm 0.2$ & $80.5 \pm 0.4$ & $63.6 \pm 0.3$ & $97.1 \pm 0.2$ & $62.8 \pm 0.2$ & $99.6 \pm 0.1$ & $80.4$ & $71.4$ \\
%\hline
RevGrad~\cite{ganin2016domain} & $79.7 \pm 0.4$ & $ 82.0 \pm 0.4$ & $68.2 \pm 0.4$ & $96.9 \pm 0.2$ & $67.4 \pm 0.5$ & $99.1 \pm 0.1$ & $82.2$ & $74.3$\\ 
%\hline
%MCD~\cite{saito2018maximum} & $74.5 \pm 0.6$ & $68.3 \pm 0.2$& $49.9 \pm 0.5$ & $90.7 \pm 0.8$ & $43.5 \pm 0.5$ & $98.3 \pm 0.5$ & $70.9 $ & $59.1$\\ 
%\hline
JAN~\cite{long2017deep} & $84.7 \pm 0.3$ & $85.4 \pm 0.3$& $68.6 \pm 0.3$ & ${ 97.4 \pm 0.2}$ & $70.0 \pm 0.4$ & ${ 99.8 \pm 0.2}$ & $84.3$ & $77.2$\\ 
%\hline
%JAN-A~\cite{long2017deep} & ${85.1 \pm 0.4}$ & ${ 86.0 \pm 0.4}$& ${ 69.2 \pm 0.4}$ & $96.7 \pm 0.3$ & ${ 70.7 \pm 0.5}$ & $99.7 \pm 0.1$ & ${ 84.6}$ & ${ 77.8}$\\  
GTA~\cite{sankaranarayanan2018generate} & ${ 87.7 \pm 0.5}$ & ${ 89.5 \pm 0.5}$& ${ 72.8 \pm 0.3}$ & $97.9 \pm 0.3$ & ${ 71.4 \pm 0.4}$ & $99.8 \pm 0.4$ & ${86.5}$ & ${80.3}$\\
%JDDA~\cite{chen2019joint} & ${ 79.8 \pm 0.1}$ & ${ 82.6 \pm 0.4}$& ${ 57.4 \pm 0.0}$ & $95.2 \pm 0.2$ & ${ 66.7 \pm 0.2}$ & $99.7 \pm 0.0$ & ${80.2}$ & ${71.6}$\\ 
%CAT~\cite{deng2019cluster} & ${90.6 \pm 1.0}$ & ${ 91.1 \pm 0.2}$& ${ 70.4 \pm 0.7}$ & $98.6 \pm 0.6$ & ${ 66.5 \pm 0.4}$ & $100 \pm 0.0$ & ${86.1}$ & ${79.7}$\\ 
CDAN~\cite{long2018conditional} & ${ 92.9 \pm 0.2}$ & ${ 94.1 \pm 0.1}$& ${ 71.0 \pm 0.3}$ & $98.6 \pm 0.1$ & ${ 69.3 \pm 0.3}$ & $100 \pm 0.0$ & ${87.7}$ & ${81.8}$\\  
\midrule
ATM [Ours]  & ${\bf 96.4 \pm 0.2}$ & ${\bf 95.7 \pm 0.3}$ & ${\bf 74.1 \pm 0.2}$ & ${\bf 99.3 \pm 0.1}$ & ${\bf 73.5 \pm 0.3}$ & ${\bf 100 \pm 0.0}$ & ${\bf 89.8}$ & {\bf 84.9}\\
boost $\uparrow$ & $\uparrow$ 3.5 & $\uparrow$ 1.6 & $\uparrow$ 3.1 & $\uparrow$ 0.7 & $\uparrow$ 4.2 & $\uparrow$ 0.0 & $\uparrow$ 2.1 & $\uparrow$ 3.1\\ 
\bottomrule
\end{tabular}
\end{table*}

\begin{table*}[ht!p]
\centering
\caption{Domain adaptation results (accuracy~\%) on Office-31. In the header, average~1 is the overall average and average~2 is the average over 4 challenging evaluations except for W$\rightarrow$D and D$\rightarrow$W. AlexNet is used for feature learning.}%The best results are highlighted by bold numbers.
\label{tab:office-alex}
\footnotesize
\vspace{-5pt}
\begin{tabular}{lcccccc|cc}
\toprule
Method & A$\rightarrow$D & A$\rightarrow$W & D$\rightarrow$A & D$\rightarrow$W & W$\rightarrow$A & W$\rightarrow$D & Avg~1 & Avg~2\\
\midrule
AlexNet~\cite{krizhevsky2012imagenet} & $63.8 \pm 0.5$ & $ 61.6 \pm 0.5$ & $51.1 \pm 0.6$ & $95.4 \pm 0.3$ & $49.8 \pm 0.4$ & $99.0 \pm 0.2$ & $70.1$ & $56.6$\\ 
DDC~\cite{tzeng2014deep} & $64.4 \pm 0.3$ & $61.8 \pm 0.4$ & $52.1 \pm 0.6$ & $95.0 \pm 0.5$ & $52.2 \pm 0.4$ & $98.5 \pm 0.4$ & $70.6$ & $57.6$ \\
%\hline
DAN~\cite{long2015learning} & $67.0 \pm 0.4$ & $68.5 \pm 0.5$ & $54.0 \pm 0.5$ & $96.0 \pm 0.3$ & $53.1 \pm 0.5$ & $99.0 \pm 0.3$ & $72.9$ & $60.7$ \\
%\hline
DRCN~\cite{ghifary2016deep} & $66.8 \pm 0.5$ & $68.7 \pm 0.3$ & $56.0 \pm 0.5$ & $96.4 \pm 0.3$ & $54.9 \pm 0.5$ & $99.0 \pm 0.2$ & $73.6$ & $61.6$ \\
%\hline
RevGrad~\cite{ganin2016domain} & $72.3 \pm 0.3$ & $ 73.0 \pm 0.5$ & $53.4 \pm 0.4$ & $96.4 \pm 0.3$ & $51.2 \pm 0.5$ & $99.2 \pm 0.3$ & $74.3$ & $62.5$\\ 
%\hline
ADDA~\cite{tzeng2017adversarial} & $71.6 \pm 0.4$ & $73.5 \pm 0.6$& $54.6 \pm 0.5$ & ${ 96.2 \pm 0.4}$ & $53.5 \pm 0.6$ & ${ 98.8 \pm 0.4}$ & $74.7$ & $63.3$\\ 
JAN~\cite{long2017deep} & $71.8 \pm 0.2$ & $74.9 \pm 0.3$& $58.3 \pm 0.3$ & ${ 96.6 \pm 0.2}$ & $55.0 \pm 0.4$ & ${ 99.5 \pm 0.2}$ & $76.0$ & $65.0$\\ 
%\hline~
%RTN~\cite{sankaranarayanan2018generate} & ${ 71.0 \pm 0.2}$ & ${ 73.3 \pm 0.3}$& ${ 50.5 \pm 0.3}$ & $96.8 \pm 0.2$ & ${ 51.0 \pm 0.1}$ & $99.6 \pm 0.1$ & ${73.7}$ & ${61.5}$\\

%MSTN\cite{long2017deep} & ${74.5 \pm 0.4}$ & ${ 80.5 \pm 0.4}$& ${ 62.5 \pm 0.4}$ & $96.9 \pm 0.1$ & ${ 60.0 \pm 0.6}$ & $99.9 \pm 0.1$ & ${ 79.1}$ & ${69.4}$\\  

CDAN~\cite{long2018conditional} & ${ 76.3 \pm 0.1}$ & ${ 78.3 \pm 0.2}$& ${ 57.3 \pm 0.2}$ & $97.2 \pm 0.1$ & ${57.3 \pm 0.3}$ & $100.0 \pm 0.0$ & ${77.7}$ & ${67.3}$\\  
\midrule
ATM [Ours]  & ${\bf 77.5 \pm 0.4}$ & ${\bf 80.2 \pm 0.3}$& ${\bf 62.1 \pm 0.5}$ & ${\bf 97.9 \pm 0.2}$ & ${\bf 61.3 \pm 0.4}$ & ${\bf 100 \pm 0.0}$ & ${\bf 79.6}$ & ${\bf 70.0}$\\ 
boost $\uparrow$ & $\uparrow$ 1.2 & $\uparrow$ 1.9 & $\uparrow$ 4.9 & $\uparrow$ 0.7 & $\uparrow$ 4.0 & $\uparrow$ 0.0 & $\uparrow$ 1.9 & $\uparrow$ 2.7 \\ 
\bottomrule
\end{tabular}
\vspace{-10pt}
\end{table*}

\subsection{Results on Digits Recognition}
The results of hand-written digits recognition are reported in Table~\ref{tab:digits}. We also report the results on source only as a baseline. The source only means the model is trained on only the source data and then directly applied on the target data. The other compared methods are mainly based on adversarial domain adaptation networks. From the results we can see that our method is able to outperform previous state-of-the-arts. Specifically, we achieve 0.5\%, 0.9\% and 6.9\% accuracy improvement on MNIST$\rightarrow$USPS, USPS$\rightarrow$MNIST and SVHN $\rightarrow$MNIST, respectively. It is worth noting that the improvement is very tough to achieve since previous state-of-the-arts are very close to the results of target full supervised setting, which is an upper bound of the performance. It can be seen that our method is quite close to the upper bound, which means that the two domains are fully aligned. At the same time, we can observe that our method achieves a significant improvement on the hardest task SVHN$\rightarrow$MNIST.

By comparing the performance of our method and CDAN on different evaluations, we can find that the advantage of our method is not quite clear on MNIST$\rightarrow$USPS and USPS$\rightarrow$MNIST. The reason is that the distribution gap between MNIST and USPS is relatively marginal. The distributions can be aligned when the domain discriminator is fully confused. However, when the distribution shift goes to large, e.g., between MNIST and SVHN, we can see that the performance of CDAN is far behind the ideal result 99.2\%, which indicates the two domains are well aligned, even when the domain discriminator has been fully confused. In this paper, we propose to simultaneously minimize the MDD metric when fooling the domain discriminator. With the guarantee that MDD is zero only if the two distributions are equal, we clearly improve the state-of-the-art performance on SVHN$\rightarrow$MNIST from 89.2\% of CDAN to 96.1\%. It is worth noting that CyCADA performs slightly better than CDAN on SVHN$\rightarrow$MNIST because it further aligns the two domains on the pixel level. CDAN and ours only align them on feature level. Anyway, our method ATM performs the best on all the three evaluations.

In Table~\ref{tab:digits}, we also report the accuracy improvements of our method over CDAN. It is worth noting that our method and CDAN share the same basic adversarial networks. The difference is that our method further introduces the proposed MDD loss. Thus, the accuracy improvements over CDAN can be seen as the contribution of the MDD. In the following tables in this paper, we will also add the ``boost $\uparrow$'' line to show the improvements, which can be seen as the results of ablation study on MDD.

\begin{table*}[ht!p]
\centering
\caption{Domain adaptation results (accuracy~\%) on ImageCLEF-DA. ResNet-50 is used for feature learning.}%The best results are highlighted by bold numbers.
\label{tab:clef}
\footnotesize
\vspace{-5pt}
\begin{tabular}{lcccccc|c}
\toprule
Method & C$\rightarrow$I & C$\rightarrow$P & I$\rightarrow$C & I$\rightarrow$P & P$\rightarrow$C & P$\rightarrow$I & ~~~Avg~~~ \\
\midrule
ResNet~\cite{he2016deep} & $78.0 \pm 0.2$ & $ 65.5 \pm 0.3$ & $91.5 \pm 0.3$ & $74.8 \pm 0.3$ & $91.2 \pm 0.3$ & $83.9 \pm 0.1$ & $80.7$ \\ 

DAN~\cite{long2015learning} & $86.3 \pm 0.4$ & $69.2 \pm 0.4$ & $92.8 \pm 0.2$ & $74.5 \pm 0.4$ & $89.8 \pm 0.4$ & $82.2 \pm 0.2$ & $82.5$  \\
%\hline
RevGrad~\cite{ganin2016domain}\!\! & $87.0 \pm 0.5$ & $ 74.3 \pm 0.5$ & $96.2 \pm 0.4$ & $75.0 \pm 0.6$ & $91.5 \pm 0.6$ & $86.0 \pm 0.3$ & $85.0$ \\ 

JAN~\cite{long2017deep} & $89.5 \pm 0.3$ & $74.2 \pm 0.3$ & $94.7 \pm 0.2$ & ${76.8 \pm 0.4}$ & $91.7 \pm 0.3$ & ${88.0 \pm 0.2}$ & $85.8$ \\ 
 
%CAT~\cite{deng2019cluster} & ${ 89.8 \pm 0.3}$ & ${ 74.0 \pm 0.2}$& ${ 94.5 \pm 0.4}$ & $76.7 \pm 0.2$ & ${ 93.7 \pm 1.0}$ & $89.0 \pm 0.7$ & ${86.3}$ \\

CDAN~\cite{long2018conditional} & ${ 91.3 \pm 0.3}$ & ${74.2 \pm 0.2}$ & ${97.7 \pm 0.3}$ & $77.7 \pm 0.3$ & ${94.3 \pm 0.3}$ & $90.7 \pm 0.2$ & ${87.7}$ \\  
\midrule
ATM [Ours]\!\!  & ~${\bf 93.5 \pm 0.1}$~ & ~${\bf 77.8 \pm 0.3}$~ & ~${\bf 98.6 \pm 0.4}$~ & ~${\bf 80.3 \pm 0.3}$~ & ~${\bf 96.7 \pm 0.2}$~ & ~${\bf 92.9 \pm 0.4}$~ & ~${\bf 90.0}$\\
boost $\uparrow$ & $\uparrow$ 2.2 & $\uparrow$ 3.6 & $\uparrow$ 0.9 & $\uparrow$ 2.6 & $\uparrow$ 2.3 & $\uparrow$ 2.2 & $\uparrow$ 2.3\\ 
\bottomrule
\end{tabular}
\end{table*}

\begin{table*}[ht!p]
\centering
\footnotesize
\caption{Domain adaptation results (accuracy~\%) on Office-Home dataset. ResNet-50 is used for feature learning.}%The best results are highlighted by bold numbers.
\label{tab:office-home}
\vspace{-5pt}
\begin{tabular}{lcccccccccccc|c}
\toprule
%\scriptsize
Method & \!A$\rightarrow$C\! & \!A$\rightarrow$P\! & \!A$\rightarrow$R\! & \!C$\rightarrow$A\! & \!C$\rightarrow$P\! & \!C$\rightarrow$R\! & \!P$\rightarrow$A\! & \!P$\rightarrow$C\! & \!P$\rightarrow$R\! & \!R$\rightarrow$A\! & \!R$\rightarrow$C\! & \!R$\rightarrow$P\! & ~Avg~ \\
\midrule
ResNet~\cite{he2016deep}  & 34.9 & 50.0 & 58.0 & 37.4 & 41.9 & 46.2 & 38.5 & 31.2 & 60.4 & 53.9 & 41.2 & 59.9 & 46.1\\ 
%\hline
DAN~\cite{long2015learning} & 43.6 & 57.0 & 67.9 & 45.8 & 56.5 & 60.4 & 44.0 & 43.6 & 67.7 & 63.1 & 51.5 & 74.3 & 56.3\\
%\hline
DANN~\cite{ganin2016domain} & 45.6 & 59.3 & 70.1 & 47.0 & 58.5 & 60.9 & 46.1 & 43.7 & 68.5 & 63.2 & 51.8 & 76.8 & 57.6\\
JAN~\cite{long2017deep} & 45.9 & 61.2 & 68.9 & 50.4 & 59.7 & 61.0 & 45.8 & 43.4 & 70.3 & 63.9 & 52.4 & 76.8 & 58.3\\
CDAN~\cite{long2018conditional} & 50.7 & 70.6 & 76.0 & 57.6 & 70.0 & 70.0 & 57.4 & 50.9 & 77.3 & 70.9 & 56.7 & 81.6 & 65.8\\ 
\midrule
ATM [Ours]  &  {\bf 52.4} & {\bf 72.6} & {\bf 78.0} & {\bf 61.1} & {\bf 72.0} & {\bf 72.6} & {\bf 59.5} & {\bf 52.0} & {\bf 79.1} & {\bf 73.3} & {\bf 58.9} & {\bf 83.4} & {\bf 67.9}\\
boost $\uparrow$ & $\uparrow$ 1.7 & $\uparrow$ 2.0 & $\uparrow$ 2.0 & $\uparrow$ 3.5 & $\uparrow$ 2.0 & $\uparrow$ 2.6 & $\uparrow$ 2.1 & $\uparrow$ 1.1 & $\uparrow$ 1.7 & $\uparrow$ 2.4 & $\uparrow$ 2.2 & $\uparrow$ 1.8 & $\uparrow$ 2.1\\ 
\bottomrule
\end{tabular}%}
\vspace{-5pt}
\end{table*}

\begin{figure*}[t]
\begin{center}
\includegraphics[width=0.98\linewidth]{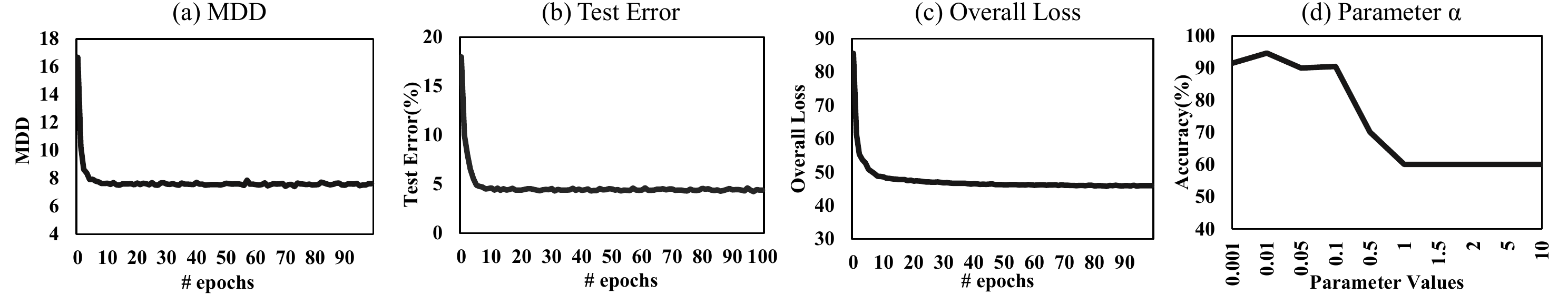}
\end{center}
\vspace{-10pt}
\caption{Model analysis on evaluation SVHN$\rightarrow$MNIST. (a) The value of MDD with different epochs. (b) The test error (\%) with different epochs. (c) The overall loss with different epochs. (d) Parameter sensitivity of $\alpha$.}
\label{fig:als1}
\vspace{-10pt}
\end{figure*}

\subsection{Objects Recognition on Office-31}
Office-31 is one of the most widely used datasets in the literatures of domain adaptation. It consists of $31$ office objects, e.g., monitor, keyboard and printer, from $3$ different domains. Samples from the DSLR~(D) domain is relatively easier to classify and samples in Amazon (A) domain is more challenging. For this dataset, we test $3\times 2=6$ evaluations in a pair-wise manner.
The results of different methods on Office-31 are reported in Table~\ref{tab:office}. In the table, we report two average results, e.g., Avg~1 and Avg~2. Avg~1 is the average over all six evaluations. Avg~2 is the average over four evaluations which excludes W$\rightarrow$D and D$\rightarrow$W since the two have very high accuracy for almost every method. As a result, Avg~2 can reflect the performance on relatively harder tasks. Comparing Avg~1 and Avg~2, we can see that most of the methods have a similar result in terms of average~1. However, the differences between average~2 are relatively larger. With the fact that the accuracy on W$\rightarrow$D and D$\rightarrow$W are close to 100\%, differences on hard tasks, e.g., D$\rightarrow$A and W$\rightarrow$A, are flooded. As a result, average~2 can reflect better the improvements. Since our method and CDAN share the similar adversarial networks, the comparison between CDAN and our ATM can reflect the effectiveness of our proposition on handling the equilibrium challenge of adversarial learning with the MDD loss. For the hardest two tasks D$\rightarrow$A and W$\rightarrow$A, our method outperforms CDAN by 3.1\% and 4.2\%, respectively. For the average performance, our proposed ATM outperforms the best baseline CDAN with 2.1\% and 3.1\% accuracy improvement in terms of Avg~1 and Avg~2, respectively.

It is worth noting that the results in Table~\ref{tab:office} are based on the features learned by ResNet-50. It can be seen that our method significantly improves the performance against CDAN. Now, one may be curious if the proposed MDD is also effective for features learned from other basic networks. To verify this, we further report the results~on~AlexNet~\cite{krizhevsky2012imagenet} in Table~\ref{tab:office-alex}. The results on AlexNet lead to the same conclusion as the results on ResNet. We can see that our method outperforms CDAN on every evaluation. Specifically,~the~performance improvements on the two hard tasks, e.g., D$\rightarrow$A and W$\rightarrow$A, are especially obvious. In terms of the average result, we outperform CDAN by 1.9\% and 2.7\% on Avg~1 and Avg~2, respectively. Combining the results in Table~\ref{tab:office} and Table~\ref{tab:office-alex}, we can observe that the proposed MDD is a generalized technique for domain adaptation. It is able to align the two domains no matter which feature representation network is~used~for~test.

\subsection{Results on ImageCLEF-DA and Office-Home}
The objects in Office-31 dataset are all from the office scenario. The resources of objects in ImageCLEF-DA dataset are more diverse, e.g., animals, vehicle and people. Therefore, we also evaluate our method on the ImageCLEF-DA dataset to verify the effectiveness in diverse scenarios. In this evaluation, we also use ResNet-50 for feature representation. The results are reported in Table~\ref{tab:clef}. From the results, we can see that when Pascal serves as the target domain, the task becomes challenging. The accuracy on other evaluations are all over 90\%. Compared with the best baseline CDAN, our method generally improves the accuracy with around 2\%, which can be seen as the contribution of our MDD. It is worth noting that the performances of the baselines on this dataset are already good. Thus, further improving the accuracy is very hard since the wrongly classified samples are very challenging. In terms of the numbers, we outperform CDAN by 2.3\% in average. The results verify that our method not only performs well on the widely used office scenario but also on more diverse scenarios.

Office-Home is a relatively large-scale domain adaptation benchmarks for classification tasks reported in recent years. Compared with Office-31, Office-Home has more categories. The results on Office-Home is reported in Table~\ref{tab:office-home}. From the results, we can see that our method performs the best on all the evaluations. Specifically, we outperform the best baseline CDAN by 2.1\% on average. Considering that Office-Home has 12 evaluations, it is remarkable to achieve the best on all of them. As a result, our proposed method not only works favorably on standard benchmarks but also generalizes well on large-scale datasets.

\subsection{Model Analysis}
\begin{figure}[t]
\begin{center}
\includegraphics[width=0.98\linewidth]{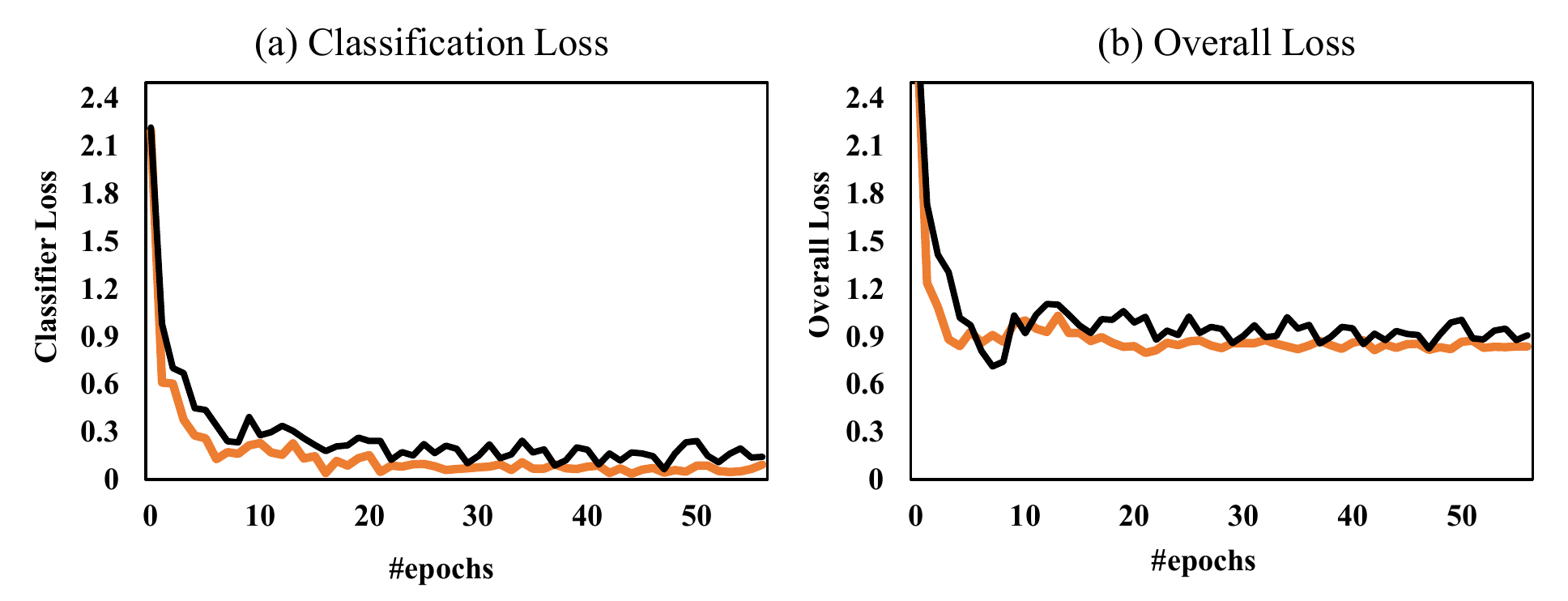}
\end{center}
\vspace{-13pt}
\caption{The classification loss (a) and overall loss (b) comparison between CDAN (black line) and our method (orange line). The evaluation A$\rightarrow$D is used as an example. Better viewed in color. It can be seen that the loss curve of our method is generally smoother than the loss curve of CDAN, which verifies that the training of our method is more stable than CDAN by further introducing the MDD divergence.} %Different colors denote different categories. 
\label{fig:loss}
\vspace{-5pt}
\end{figure}

{\bf Training Stability.} The adversarial networks are generally known as hard to train. In Fig.~\ref{fig:als1}(b) and Fig.~\ref{fig:als1}(c) we report the test error and overall loss with different epochs on SVHN$\rightarrow$MNIST. It can be seen that our method is able to converge within 20 epochs on SVHN$\rightarrow$MNIST. At the same time, we can also observe that the test error and the overall loss have very similar trends, which means the training is stable and effective. It is worth noting that we only calculate the inter-domain distance at the corresponding positions instead of calculating all the pair distances in Eq.~\eqref{eq:mddloss}. We claimed that such a paradigm is able to lead to convergence and reduce the computational complexity. The result in Fig.~\ref{fig:als1}(b) and Fig.~\ref{fig:als1}(c) verify this claim. In Fig.~\ref{fig:loss}, we report more results on the evaluation A$\rightarrow$D to analyze the equilibrium issue. The results also verify that optimizing the MDD in a batch instead of the whole dataset can also achieve convergence.

\begin{figure*}[t!h]
\begin{center}
\subfigure[Original]{
\includegraphics[width=0.3\linewidth]{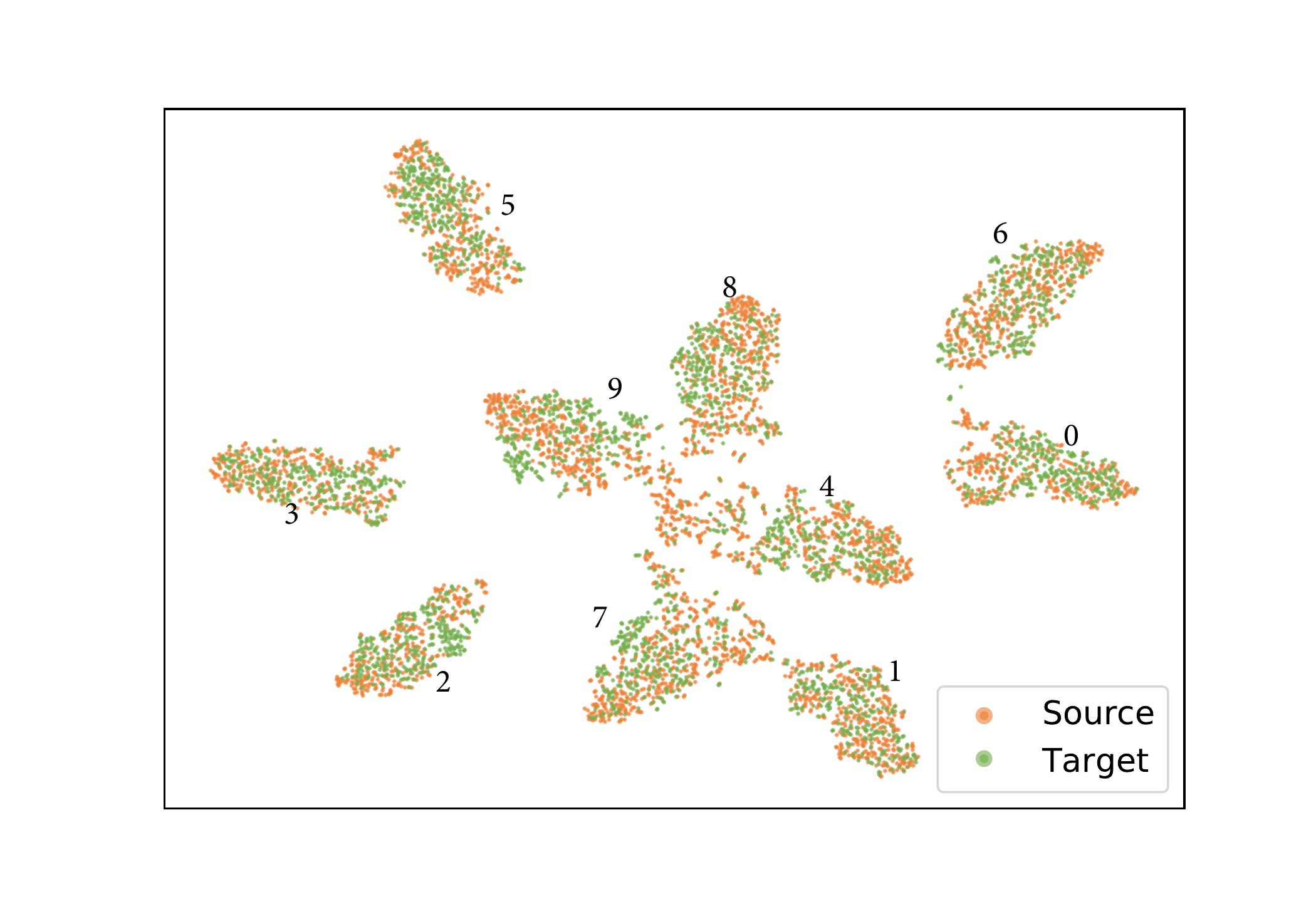}
}
\subfigure[CDAN]{
\includegraphics[width=0.3\linewidth]{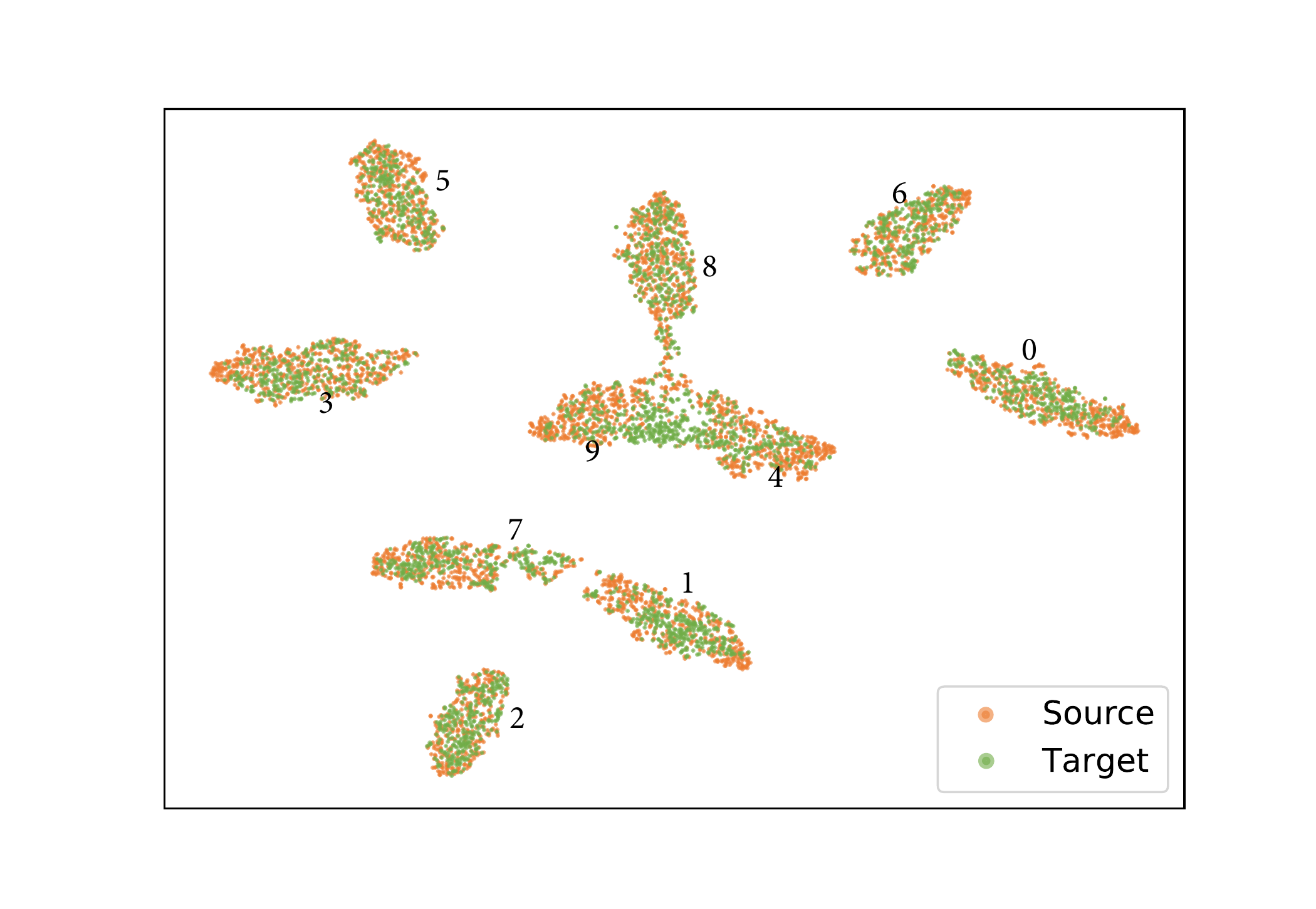}
}
\subfigure[Ours]{
\includegraphics[width=0.3\linewidth]{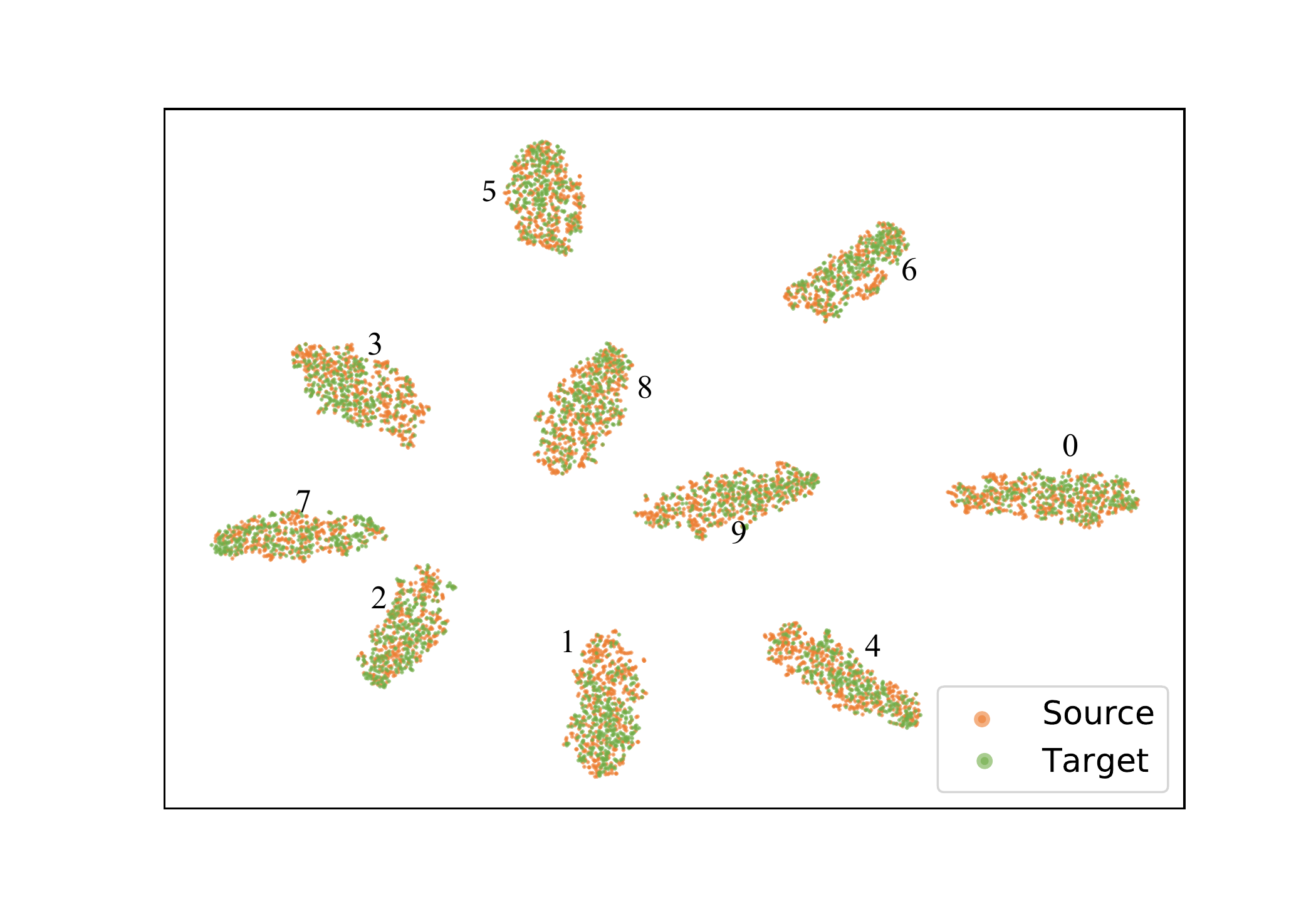}
}
\end{center}
\vspace{-12pt}
\caption{Visualization of the learned representations by t-SNE~\cite{maaten2008visualizing}. The evaluation SVHN$\rightarrow$MNIST is used as an example. Specifically, figure (a), (b) and (c) visualize the original representations (non-adapted), CDAN and our representations, respectively. The number near each cluster is the corresponding category label. It can be seen that some classes, e.g., 4 and 9, are still confusing in CDAN. Our method has good transferability and discriminability with the power of MDD.}
\label{fig:vis}
\vspace{-12pt}
\end{figure*}

\begin{figure}[t]
\begin{center}
\includegraphics[width=0.98\linewidth]{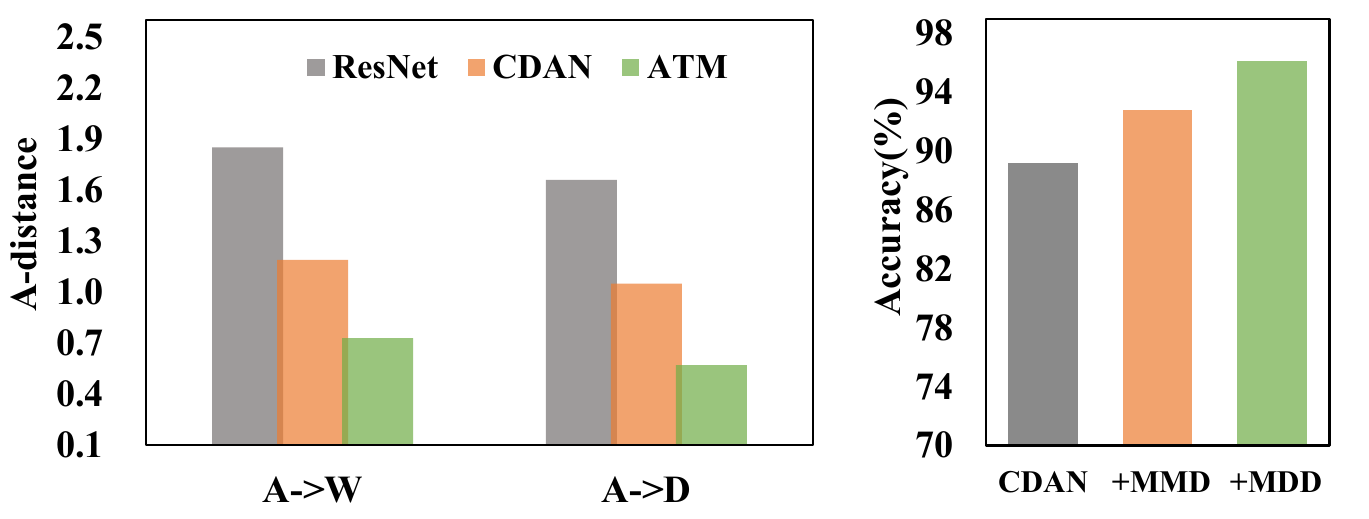}
\end{center}
\vspace{-15pt}
\caption{The effectiveness of the MDD. The left sub-figure reports the $\mathcal{A}$-distance of different methods on A$\rightarrow$D. The right sub-figure shows the results of different settings, i.e., CDAN, CDAN+MMD and CDAN+MDD, on SVHN$\rightarrow$MNIST.} %Different colors denote different categories. 
\label{fig:als2}
\vspace{-10pt}
\end{figure}

{\bf Parameter Sensitivity.} In our model, the hyper-parameter is $\alpha$ which controls the loss of MDD. To better understand the effect of $\alpha$, we report the sensitivity of $\alpha$ in Fig.~\ref{fig:als1}(d). It can be seen that our ATM achieves the best result with $\alpha=0.01$. As a result, we fix $\alpha=0.01$ in this paper. One can tune the parameter by importance-weighted cross-validation for their own applications. Here, it is worth noting that $0.01$ is a relatively small number. Thus, one may doubt the contribution of MDD with such a small weight. To address this, we can go back to review the adversarial loss and the MDD loss. As we can observe, the adversarial loss has a {\it log} operation on the output of the domain discriminator. However, the MDD loss is the sum of mean squared $\ell_2$ distance. As a result, it is not hard to speculate that the absolute value of the MDD loss could be much larger than the adversarial loss. Therefore, a small weight on the MDD loss ensures that it would be reweighed to the same scale of the adversarial loss.

{\bf The Effectiveness of MDD.} Our ATM shares the similar adversarial network with CDAN. The difference is that we further optimize MDD to alleviate the equilibrium challenge. By comparing the results of our ATM and CDAN, it can be seen from Table~\ref{tab:digits}-Table~\ref{tab:office-home} that our MDD is effective in handling domain adaptation problems. In Fig~\ref{fig:als1}(a), we further report the MDD with different epochs. The figure reflects that MDD is stably reduced with the iterations until convergence. Combining the results in Fig~\ref{fig:als1}(a)-(c), we can see that the MDD loss and the adversarial loss serve the same purpose for aligning the data distributions.

In addition, to verify the effectiveness on handling the equilibrium issue, we report the classification loss and overall loss of both CDAN and our method in Fig.~\ref{fig:loss}. It is easy to see that the loss curve of our method is smoother, which indicates that the equilibrium between the generator and the discriminator is easier to be achieved in our method. We can also observe that the classification loss of our method is consistently lower than CDAN, which implies that the representations learned by our method are more effective. At last, it is worth noting that MDD loss is nonnegative. With the additional nonnegative MDD, our method is still better towards the convergence.

The left sub-figure in Fig.~\ref{fig:als2} shows the $\mathcal{A}$-distance of our method and CDAN. $\mathcal{A}$-distance is a widely used metric to measure the distribution divergence~\cite{ben2010theory}. The $\mathcal{A}$-distance is defined as $\mathcal{A}_\mathrm{dis}=2(1-2\epsilon)$, where $\epsilon$ denotes the test error of a classifier which is trained to distinguish the source and target domains. The smaller the $\mathcal{A}$-distance, the better the distribution alignment. It can be seen that our method is able to achieve smaller $\mathcal{A}$-distance than CDAN, which means our method is better at aligning the two domains. The right sub-figure in Fig.~\ref{fig:als2} reports the adaptation results of CDAN+MMD and CDAN+MDD on the evaluation SVHN$\rightarrow$MNIST, we can see that our MDD achieves better result than MMD.

{\bf Ablation Study.} Our method mainly consists of two parts: the adversarial training and the MDD minimization. If we remove the MDD part, our method would be similar to CDAN, of which the results have been reported. On the other hand, if we only deploy the MDD loss on the feature learner $F$ to extract features, we achieve an accuracy of 73.9\% on SVHN$\rightarrow$MNIST which is better than the baseline source only, DAN with MMD metric and DANN with H-divergence.

Furthermore, since our MDD consists of 3 terms as reported in~Eq.~\eqref{eq:bddistance}. We report the intra-MDD ablation study in Table~\ref{tab:ablation} to show the contribution of each term. From the results, we can see that each term in MDD has specific contribution. The three working together can achieve the best performance.

\begin{table}[t]
\centering
\footnotesize
\caption{Intra-MDD ablation study on evaluation S$\rightarrow$M. ``T1'' is short for ``Test~1'', ``$\bullet$'' indicates the term is used. Term~1 denotes the first term in Eq.~\eqref{eq:bddistance}, and so on.}
\label{tab:ablation}
\vspace{-5pt}
\begin{tabular}{lcccccccc}
\toprule
%\tabucline[0.75pt]{-----}
Settings & T1 & T2 & T3 & T4 &T5 &T6 &T7 &T8 \\
\midrule
Term~1 & $\circ$ &$\bullet$ & $\circ$     & $\circ$     & $\bullet$ & $\bullet$ &$\circ$     &$\bullet$ \\
Term~2 & $\circ$ &$\circ$     & $\bullet$ & $\circ$     & $\bullet$ & $\circ$     &$\bullet$ &$\bullet$\\
%\hline
Term~3 & $\circ$ &$\circ$     & $\circ$     & $\bullet$ & $\circ$     &$\bullet$  &$\bullet$ &$\bullet$\\
\midrule
Result& $89.2$ & 93.4 & 92.6 & 91.9 & 95.3 & 95.0 & 93.8 & 96.1 \\
\bottomrule
\end{tabular}%}
\vspace{-10pt}
\end{table}

\begin{figure}[t]
\begin{center}
\includegraphics[width=0.9\linewidth]{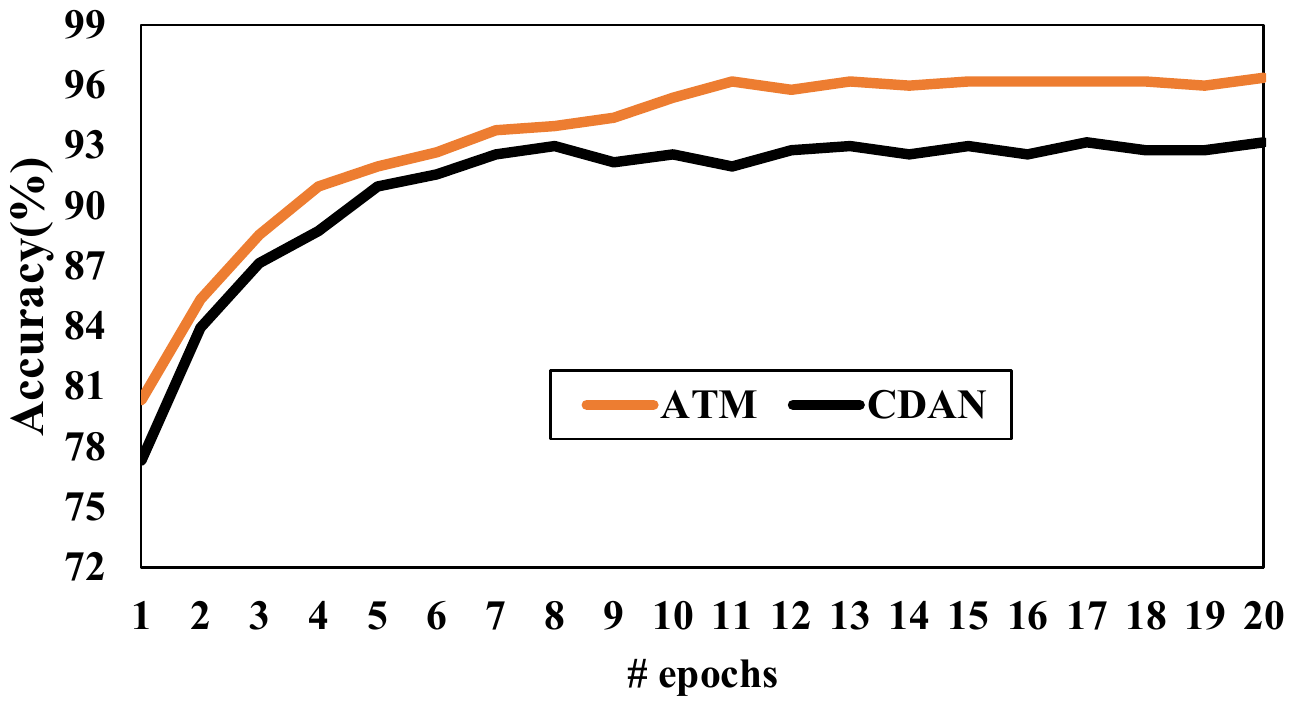}
\end{center}
\vspace{-15pt}
\caption{Results of pseudo labeling on A$\rightarrow$D.} %Different colors denote different categories. 
\label{fig:pseudo}
\vspace{-10pt}
\end{figure}

{\bf Pseudo Labeling.} To take care of the conditional distribution, we deploy pseudo labeling to get the class information of the target domain samples. It is worth noting that the pseudo labeling is not fixed as in initialization. The pseudo labels are dynamically updated in each iteration. Fig.~\ref{fig:pseudo} reports the accuracy of the pseudo labeling on A$\rightarrow$D in different iterations. It can be seen that the accuracy of pseudo labeling is steadily increased with the iterations.

{\bf Visualization.} Fig.~\ref{fig:vis} reports the t-SNE~\cite{maaten2008visualizing} visualization of the features learned by different methods. The figure vividly shows that our method can learn discriminative features for the target domain data. It also verifies that our model is capable of aligning the two data distributions.

\begin{figure}[t]
\begin{center}
\includegraphics[width=0.9\linewidth]{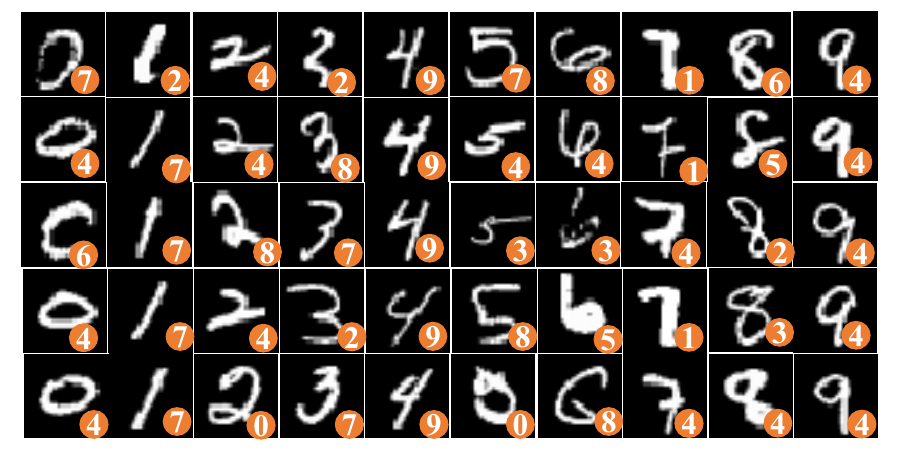}
\end{center}
\vspace{-10pt}
\caption{Sample review. This figure shows the randomly selected samples from each class which are wrongly classified by CDAN but correctly classified by our ATM. The small numbers in the bottom right corner denote the labels predicted by CDAN. It can be seen that some images, i.e, the images of 4 and 9, are easy to confuse the domain discriminator and then lead to the wrong results of CDAN. Our method, however, is able to alleviate such issues by leveraging the proposed MDD loss.} %Different colors denote different categories. 
\label{fig:sample1}
\vspace{-5pt}
\end{figure}

{\bf Qualitative Study.} Apart from the quantitative results, Fig.~\ref{fig:sample1} shows some randomly selected samples which are wrongly classified by CDAN but correctly categorized by our ATM. It can be seen that our method is more robust. It learns better feature representations by aligning the two domains. It is able to classify some very confusing categories, e.g., handwritten $4$ and $9$.

\section{Conclusion}
In this paper, we propose a novel method Adversarial Tight Match (ATM) for unsupervised domain adaptation. We verify that the equilibrium challenge of adversarial learning in domain adaptation can be alleviated by jointly optimizing an additional loss which measures the distribution divergence between the two domains. Furthermore, we propose a novel loss Maximum Density Divergence (MDD) which manages to align the two domains by simultaneously minimizing the inter-domain divergence and maximizing the intra-class density. Specifically, MDD is tailored to be smoothly incorporated into the adversarial domain adaptation framework. Extensive experiments on four benchmarks verify that our ATM can outperform previous state-of-the-arts with significant advantages. In our future work, we will investigate the more challenging~domain~generalization~problem.% The proposed method presents a novel solution for challenging equilibrium issue in adversarial training.

\ifCLASSOPTIONcaptionsoff
  \newpage
\fi

\bibliographystyle{IEEEtran}
\bibliography{mm19}

%\vspace{-60pt}
\begin{IEEEbiography}[{\includegraphics[width=1in,height=1.25in,clip,keepaspectratio]{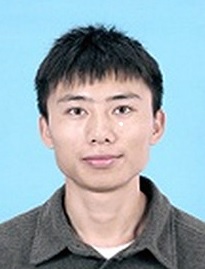}}]{Jingjing Li}
 received his MSc and PhD degree in Computer Science from University of Electronic Science and Technology of China in 2013 and 2017, respectively. Now he is an associate professor with the School of Computer Science and Engineering, University of Electronic Science and Technology of China. He has great interest in machine learning, especially transfer learning, zero-shot learning and recommender systems. 
 \end{IEEEbiography}

 \vspace{-50pt}
 \begin{IEEEbiography}[{\includegraphics[width=1in,height=1.25in,clip,keepaspectratio]{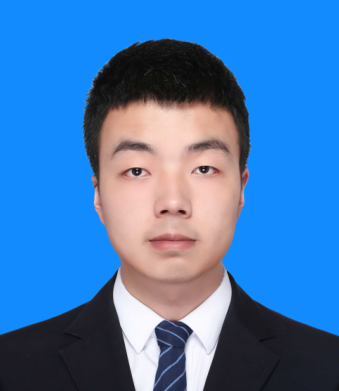}}]{Erpeng Chen}
 received his B.Sc. degree in 2016 from University of Electronic Science and Technology. He is currently a second year Master student in the University of Electronic Science and Technology of China. His research interests include machine learning and computer vision.
 \end{IEEEbiography}
 \vspace{-50pt}

 \begin{IEEEbiography}[{\includegraphics[width=0.9in,height=1.25in,clip,keepaspectratio]{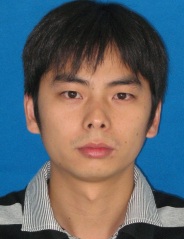}}]{Zhengming Ding}
 received the B.Eng. and the M.Eng. degree from University of Electronic Science and Technology of China (UESTC), China, in 2010 and 2013, respectively. He received the Ph.D. degree from the Department of Electrical and Computer Engineering, Northeastern University, USA in 2018. He is a faculty member affiliated with Department of Computer, Information and Technology, Indiana University-Purdue University Indianapolis since 2018. %He is currently an Associate Editor of the Journal of Electronic Imaging (JEI).
\end{IEEEbiography}
 \vspace{-50pt}

\begin{IEEEbiography}[{\includegraphics[width=1in,height=1.25in,clip,keepaspectratio]{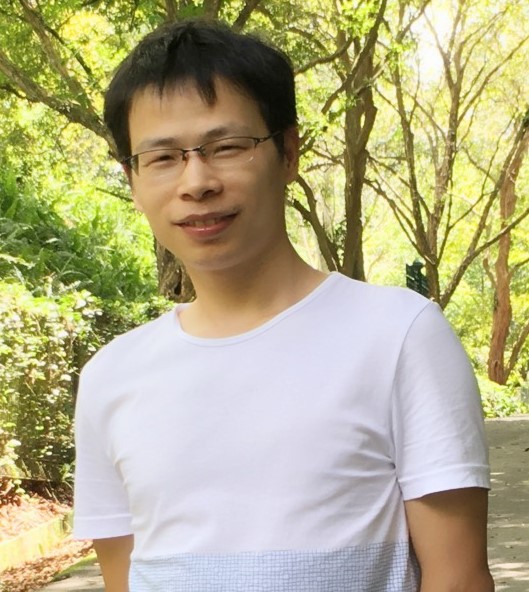}}]{Lei Zhu} received the B.S. degree (2009) at Wuhan University of Technology, the Ph.D. degree (2015) at Huazhong University of Science and Technology. He is currently a full Professor with the School of Information Science and Engineering, Shandong Normal University, China. He was a Research Fellow under the supervision of Prof. Heng Tao Shen at the University of Queensland (2016-2017), and Dr. Jialie Shen at the Singapore Management University (2015-2016). His research interests are in the area of large-scale multimedia content analysis and retrieval.
\end{IEEEbiography}
\vspace{-50pt}

\begin{IEEEbiography}[{\includegraphics[width=1in,height=1.25in,clip,keepaspectratio]{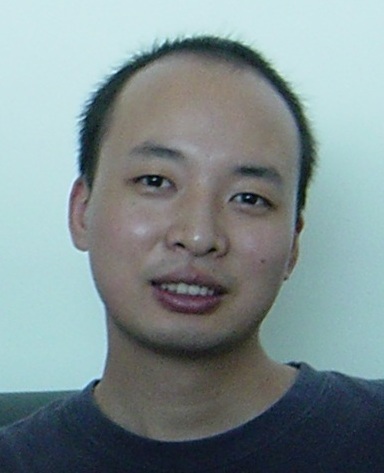}}]{Ke Lu}
 received his B.Sc. degree from Chongqing University, China in 1996. He obtained his MSc and PhD degrees in Computer Application Technology from the University of Electronic Science and Technology of China, in 2003 and 2006, respectively. He is currently a professor in School of Computer Science and Engineering, University of Electronic Science and Technology of China. His research interests include pattern recognition, multimedia~and~computer~vision.
 \end{IEEEbiography}
 \vspace{-50pt}

 \begin{IEEEbiography}[{\includegraphics[width=1in,height=1.25in,clip,keepaspectratio]{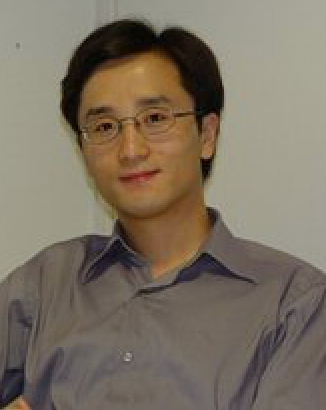}}]{Heng Tao Shen} is currently a Professor of National "Thousand Talents Plan", the Dean of School of Computer Science and Engineering, and the Director of Center for Future Media at the University of Electronic Science and Technology of China. He is also an Honorary Professor at the University of Queensland. He obtained his BSc with 1st class Honours and PhD from Department of Computer Science, National University of Singapore in 2000 and 2004 respectively. His research interests mainly include Multimedia Search, Computer Vision, Artificial Intelligence, and Big Data Management. He currently is an Associate Editor of IEEE Transactions on Knowledge and Data Engineering and IEEE Transactions on Image Processing. He is an ACM Distinguished Member and an OSA Fellow.
\end{IEEEbiography}


%\vfill


\end{document}